\documentclass{article} 
\usepackage{iclr2027_preprint,times}

\usepackage{amsmath,amsfonts,bm}

\def\eqref#1{equation~\ref{#1}}

\def\1{\bm{1}}

\DeclareMathAlphabet{\mathsfit}{\encodingdefault}{\sfdefault}{m}{sl}
\SetMathAlphabet{\mathsfit}{bold}{\encodingdefault}{\sfdefault}{bx}{n}

\usepackage[utf8]{inputenc} 
\usepackage[T1]{fontenc}    
\usepackage{hyperref}       
\usepackage{url}            
\usepackage{booktabs}       
\usepackage{amsfonts}       
\usepackage{nicefrac}       
\usepackage{microtype}      
\usepackage[table]{xcolor}   
\usepackage{graphicx}
\usepackage{subcaption}
\usepackage{caption}
\usepackage{capt-of}
\usepackage[textsize=tiny]{todonotes}
\usepackage[most]{tcolorbox}
\usepackage{enumitem}
\usepackage{xspace,soul}
\usepackage{duckuments}
\usepackage{lipsum}
\usepackage{multirow}
\definecolor{xblue}{HTML}{4169E1}      
\definecolor{xgreen}{HTML}{036C3A}
\definecolor{xpurple}{HTML}{9838B1}    
\definecolor{xslategray}{HTML}{70818F} 
\definecolor{xorange}{HTML}{FF8C00}    
\definecolor{xcyan}{HTML}{06AEEF}
\definecolor{xred}{HTML}{FF0000}       
\definecolor{xgray}{HTML}{808080}
\definecolor{xxgreen}{HTML}{009F86}    
\definecolor{xsienna}{HTML}{8B4512}
\definecolor{xxpurple}{HTML}{623E99}
\definecolor{xorchid}{HTML}{BE6BD6}    
\definecolor{xindigo}{HTML}{472B7A}    

\newcommand{\xslategray}[1]{\textcolor{xslategray}{#1}}

\usepackage{algorithm}
\usepackage[noend]{algpseudocode}

\algrenewcommand{\algorithmiccomment}[1]{\hfill \xslategray{\texttt{\#~#1}}}
\usepackage[most]{tcolorbox}
\usepackage{wrapfig}

\usepackage{fvextra}
\usepackage{xcolor}

\usepackage[most]{tcolorbox}

\definecolor{propback}{RGB}{245,247,250}
\definecolor{propframe}{RGB}{130,142,160}

\newtcolorbox{propositionbox}[1]{
    colback=propback,
    colframe=propframe,
    boxrule=0.45pt,
    arc=1mm,
    left=1.5mm,
    right=1.5mm,
    top=1.2mm,
    bottom=1.2mm,
    before skip=6pt,
    after skip=6pt,
    title={\textbf{Proposition 1.} #1},
    fonttitle=\normalfont
}

\title{Seeing Is Not Addressing: Auditing linguistic Access to Frozen Visual Geometry}

\author{
\makebox[0.75\textwidth][s]{%
Woosang Jeon$^{1,}$\thanks{Equal contribution.}
\hspace{0.2em}
\hfill
Jiwon Yang$^{1,*}$
\hspace{-0.4em}
\hfill
Soo Chung$^{1}$
\hspace{-0.4em}
\hfill
Taehyeong Kim$^{1,}$\thanks{Corresponding author.}%
} \\
$^{1}$Seoul National University \\
\texttt{\{jwoosang1,jwyang0424,soochung,taehyeong.kim\}@snu.ac.kr}
}
\iclrfinalcopy 
\begin{document}

\maketitle

\begin{abstract}
Visual distinctions are often finer than those reflected in linguistic conceptualization.
Vision--language models exhibit a similar asymmetry: a distinction can remain discriminable in frozen image geometry while being weakly addressable through the native text interface.
We study this gap by separating \textit{visual discriminability} from \textit{linguistic addressability} in text-to-image retrieval.
Using FactorAtlas, a fully crossed testbed of 23,040 images spanning shape, hue, pattern, and nuisance variation, we compare both readouts on held-out images of the same distinctions.
We then derive image-side contrasts that separate each value from its alternatives for matched visual grounding, and test whether this reduces the native-text access gap across factors and models.
Direction-specific and visual-absence controls tie these gains to the relevant visual contrast; the gains persist after global alignment and extend to compositional retrieval and natural images.
Together, these results show that visual discriminability and linguistic addressability need not coincide, and that matched visual grounding can probe and reduce the resulting access gap.
\end{abstract}


\section{Introduction}

Visual representations often preserve finer distinctions than are reflected in linguistic conceptualization~\citep{liao2024probing}.
Such distinctions become addressable through language when relevant words are linked to their visual referents, and these associations are learned through repeated co-occurrence~\citep{yu2007rapid,smith2008infants}.
Contrastively trained vision--language models have a similar learning structure, acquiring cross-modal correspondences from large-scale paired image--text data~\citep{radford2021learning,zhai2023sigmoid}.
The paired text, however, need not describe every visual distinction present in the image.
Accordingly, in the resulting shared embedding space, some of these distinctions can remain \textbf{discriminable in the image representation while being only weakly addressable} through the model's text interface.

In \textbf{text-to-image retrieval}, this asymmetry motivates separating two aspects of performance: whether image geometry supports the relevant visual distinction, and how well the native text query can access it.
Related work shows that representations within each modality retain structured information~\citep{koishigarina2026clip}, with compositional and ordinal structure in image embeddings~\citep{berasi2025not,sonthalia2026rankability}.
Meanwhile, cross-modal alignment and mismatch have been studied at broader representational and compositional scales~\citep{liang2022mind,kamath2023text,koishigarina2026clip}.
We instead take \textbf{each named visual distinction as the unit of analysis}, asking how the image-side structure underlying its discriminability relates to native linguistic access.

We operationalize this question using \textbf{FactorAtlas}, a fully crossed testbed of 23,040 images spanning 8 shapes, 12 hues, 10 patterns, and 24 nuisance realizations.
For each named factor value, we compare two readouts under held-out conditions: \textbf{\textit{visual discriminability}}---how reliably each value can be distinguished from its alternatives in frozen image geometry---and \textbf{\textit{linguistic addressability}}---how well the corresponding text query retrieves images exhibiting that value.
We then derive a target-versus-rest visual contrast for each value and use its direction to ground the native text query.
We call this \textbf{\textit{matched visual grounding}} and test whether it improves distinction-specific linguistic access.

Across our experiments, visual discriminability and linguistic addressability diverge, while matched visual grounding yields held-out access gains across factors and models.
These gains are direction-specific and weaken as the underlying visual contrast is attenuated. 
Broad alignment reduces part of the mismatch but leaves residual value-specific access gaps that matched grounding further reduces. 

These access gains, in turn, support exact compositional retrieval under held-out conditions, while the same value-specific intervention can be applied selectively and generalizes to natural images.
Together, these findings show that linguistic access can lag behind visual structure already present in frozen representations, suggesting a way to expose and use under-accessed visual distinctions.


\section{From Visual Distinctions to Linguistic Access}
\label{sec:audit}

\subsection{Visual Discriminability and Linguistic Addressability}
\label{sec:readouts}

Our analysis separates two questions for the same named visual property: whether its distinction remains discriminable in frozen image geometry and how well the native text query can access it.
We term these \textit{visual discriminability} and \textit{linguistic addressability}, measured from image-derived evidence and the native text query, respectively.
When they diverge, we use the held-out improvement obtained by grounding the native text query with the corresponding image-side structure, which we call \textit{matched access gain}, as an interventional readout of how much the access gap can be reduced.

\begin{figure}[h]
    \centering
    \includegraphics[
        width=0.95\linewidth,
        trim={0.2cm 0.2cm 0.2cm 0.2cm},
        clip
    ]{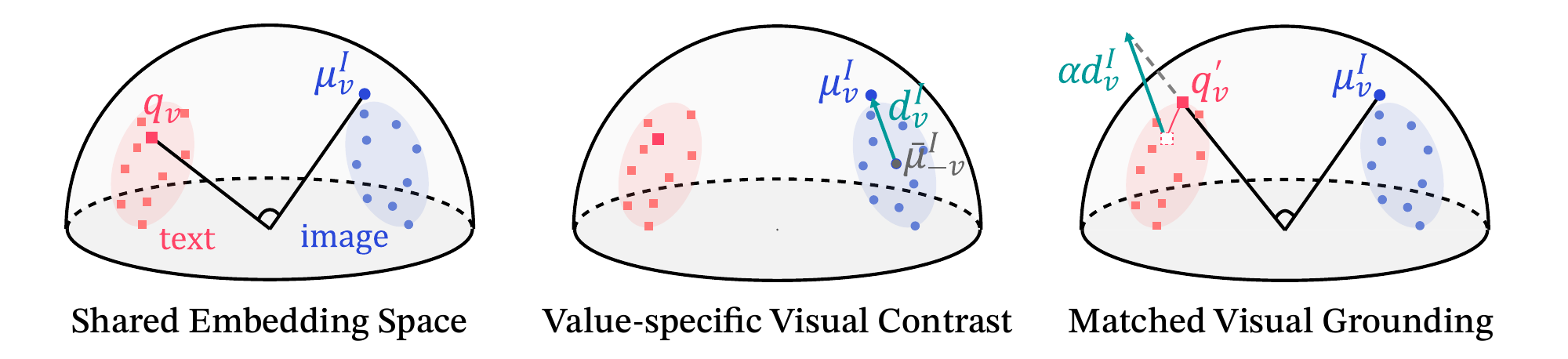}
    \vspace{-0.6em}
    \caption{
    \textbf{From visual discriminability to linguistic access.}
    A visual distinction may remain discriminable in frozen image geometry yet weakly accessed by its native text query; matched visual grounding connects the two through a value-specific visual contrast.
    }
    \label{fig:main_concept}
\end{figure}

\subsection{Matched Visual Grounding via Value-Specific Visual Contrasts}
\label{sec:grounding}

We instantiate this concept by representing each visual value $v$ with an image-side direction that distinguishes it from the other values in the same factor vocabulary.
We use this direction to ground the corresponding native text query, a procedure we call \textit{matched visual grounding} (Fig.~\ref{fig:main_concept}).

Let $I(x)\in\mathbb{R}^d$ denote the frozen image embedding.
For a factor with value set $\mathcal{V}$ and value $v\in\mathcal{V}$, let $C_v$ denote a visual support set for $v$.
We define the target and average-rest image prototypes as
\[
\mu_v^I
=
\operatorname{norm}\!\left(
\frac{1}{|C_v|}
\sum_{x\in C_v}
\operatorname{norm}(I(x))
\right),
\qquad
\bar{\mu}_{-v}^I
=
\frac{1}{|\mathcal{V}|-1}
\sum_{u\neq v}\mu_u^I.
\]
We leave $\bar{\mu}_{-v}^I$ unnormalized so that its score under a unit query equals the average score of the remaining value prototypes.
Their difference defines the matched visual contrast and its unit direction:
\[
\Delta_v^I
=
\mu_v^I-\bar{\mu}_{-v}^I,
\qquad
d_v^I
=
\frac{\Delta_v^I}{\|\Delta_v^I\|}.
\]
Given the native text query $q_v$, we ground it along this direction:
\begin{equation}
q_v'(\alpha)
=
\operatorname{norm}\!\left(
q_v+\alpha d_v^I
\right),
\qquad
\alpha\geq0.
\label{eq:correction}
\end{equation}
The parameter $\alpha$ controls the grounding strength.
Comparing retrieval with $q_v'(\alpha)$ against the native query $q_v$ yields the \textit{matched access gain} introduced above.

\subsection{Geometric Guarantee}
\label{sec:guarantee}

We derive a guarantee in terms of the target-versus-average-rest similarity margin.
\[
M_v(q)=q^\top\Delta_v^I,
\qquad
c=q_v^\top d_v^I.
\]

\begin{propositionbox}{Matched similarity-margin guarantee}
For every $\alpha\geq0$ for which $q_v'(\alpha)$ is defined,
\[
M_v\!\left(q_v'(\alpha)\right)\geq M_v(q_v).
\]
\end{propositionbox}

The grounded similarity margin has the closed form
\begin{equation}
M_v\!\left(q_v'(\alpha)\right)
=
\|\Delta_v^I\|
\frac{c+\alpha}
{\sqrt{1+\alpha^2+2\alpha c}}.
\end{equation}

For $|c|<1$, $M_v(q_v'(\alpha))$ increases strictly with $\alpha$.
When the initial margin is negative ($-1<c<0$), it crosses zero at
\[
\alpha^\star=-c=-q_v^\top d_v^I.
\]

The guarantee is local to the matched target-versus-average-rest margin; whether it translates into held-out retrieval gains and broader retrieval behavior is evaluated empirically below. 
The full proof, including endpoint cases, is given in Appendix~\ref{app:proof}.


\section{Diagnosing and Reducing Distinction-Level Access Gaps}
\label{sec:evidence}

\begin{figure}[h]
    \centering
    \includegraphics[
        width=\linewidth,
        trim={1.0cm 1.0cm 0.8cm 0.5cm},
        clip
    ]{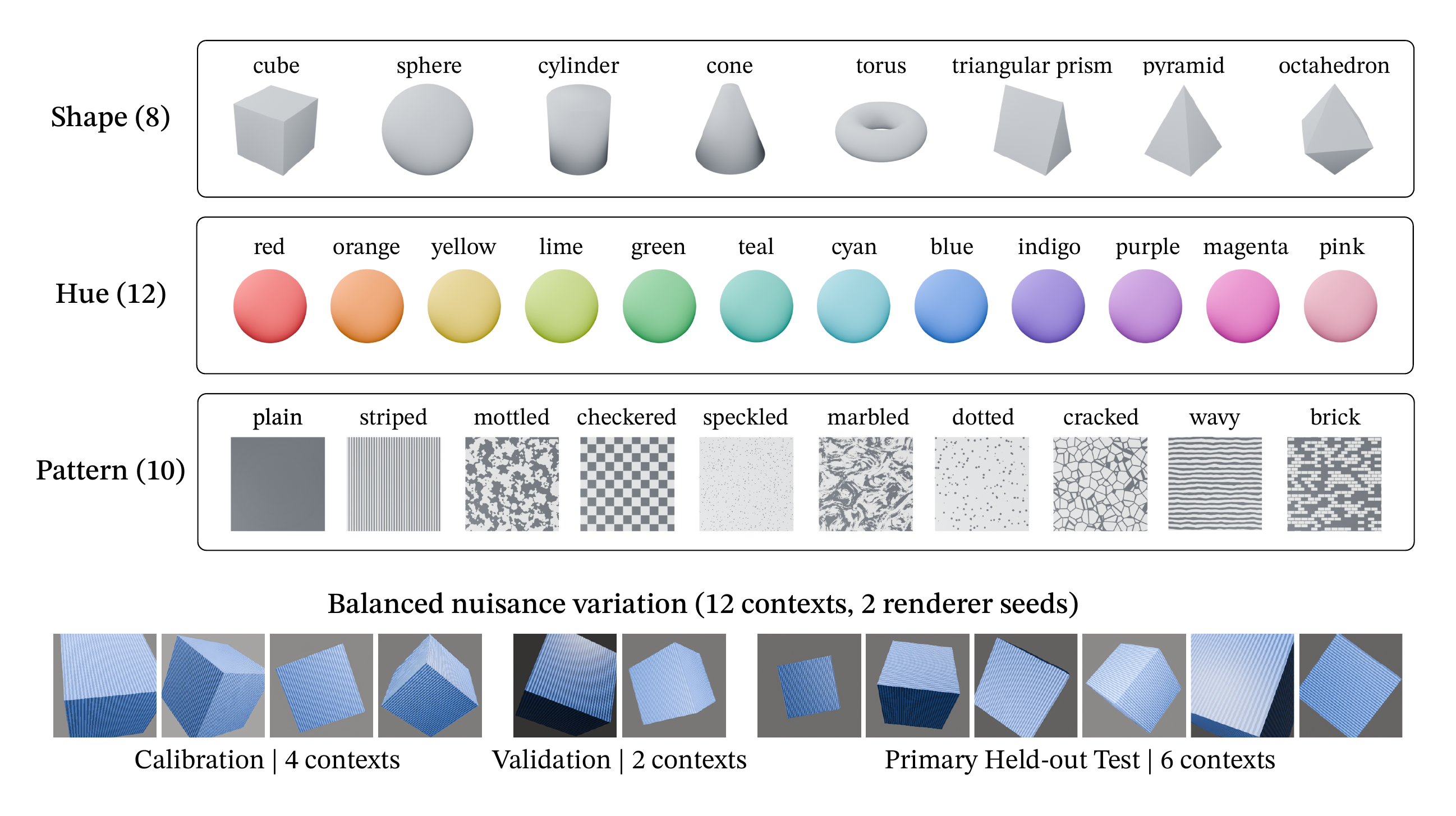}
    \caption{
    \textbf{FactorAtlas and its primary held-out protocol.}
    The testbed factorially combines 8 shapes, 12 hues, and 10 surface patterns under balanced nuisance variation.
    Disjoint nuisance contexts are used for calibration, validation, and held-out testing, while the semantic factors remain fully crossed and balanced.
    }
    \label{fig:factoratlas}
\end{figure}

\subsection{FactorAtlas and Evaluation Protocol}
\label{sec:factoratlas}

FactorAtlas is a procedurally rendered visual testbed designed to compare visual discriminability and linguistic addressability under controlled variation.
It contains 23,040 images spanning 8 shapes, 12 hues, 10 surface patterns, and 24 nuisance realizations.
Each realization is jointly determined by a context and renderer seed, varying object pose, viewpoint, illumination, background, and material appearance while preserving the underlying semantic factors.
This lets the same semantic distinctions recur across varied visual conditions.

For each semantic factor, we evaluate all of its values jointly. 
When one factor is the target, the other factors are fully crossed and balanced, so every target value appears equally often with every combination of the remaining factors. 
This rules out simple cross-factor frequency shortcuts, preventing the readout from relying on correlations between the target value and the other factors.

Our primary protocol holds out nuisance contexts, using four for calibration, two for validation, and six for testing, with both renderer seeds of each context assigned to the same split.
For each factor value, the native text query is formed by averaging and renormalizing the embeddings of three fixed prompt templates specific to its factor.
Calibration defines the visual support sets in Section~\ref{sec:grounding}, while validation selects the factor-level grounding strength $\alpha$.
All choices are fixed before evaluation on held-out contexts, so the readouts are tested under nuisance conditions not used to construct or tune them.
Full construction and evaluation details are provided in Appendix~\ref{app:factoratlas_protocol}.

\subsection{Discriminability, Addressability, and Matched Access}
\label{sec:canonical}

\paragraph{Discriminability versus addressability.}

\begin{wrapfigure}[12]{r}{0.42\textwidth}
    \centering
    \includegraphics[
        width=0.38\textwidth,
        trim={0.4cm 0.4cm 0.4cm 0.8cm}
    ]{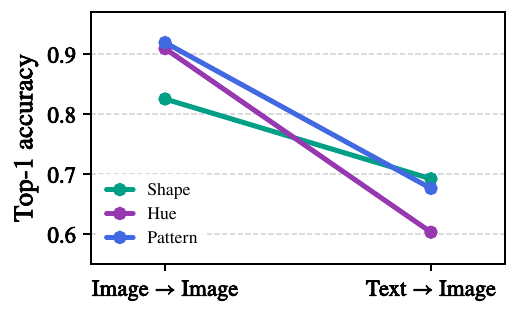}
    \caption{
    \textbf{Factor-value top-1 readout gap.}
    Visual discriminability exceeds native linguistic addressability across all factors.
    }
    \label{fig:factor_level_separation}
\end{wrapfigure}

We first compare \textit{visual discriminability} and \textit{linguistic addressability} on the same held-out gallery.
For each factor value, the former is measured with an image-derived query built from calibration images, whereas the latter uses the corresponding native text query.
Thus, the two readouts differ only in how the query is obtained.

As shown in Fig.~\ref{fig:factor_level_separation}, image-based queries identify the correct value more reliably than native text queries across shape, hue, and pattern.
This reveals the distinction-level asymmetry motivating our audit, as the relevant visual distinctions remain readily discriminable in image geometry even when their linguistic handles provide weaker access.

\begin{figure}[h]
    \centering
    \includegraphics[
        width=0.98\linewidth,
        trim={0.2cm 0.3cm 0.2cm 0.2cm},
        clip
    ]{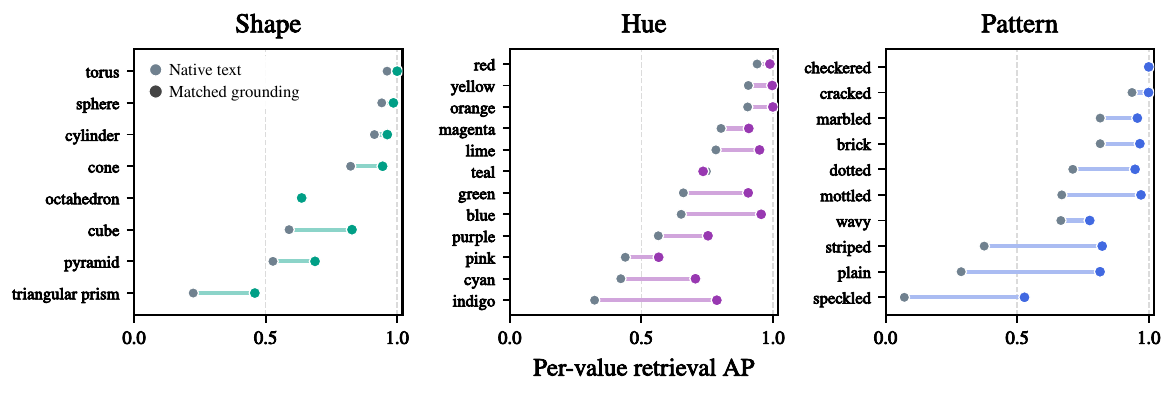}
    \caption{
    \textbf{Matched access gain across individual factor values.}
    Native text retrieval and matched visual grounding are compared for each value of shape, hue, and pattern on held-out images.
    }
    \label{fig:per_value_recovery}
\end{figure}

\paragraph{Matched access gain across factors.}

We next ask whether grounding the text query with value-specific image-side information can reduce this access gap.
Using the matched visual grounding procedure in Section~\ref{sec:grounding}, we estimate each visual direction from calibration images and select the factor-level strength $\alpha$ on validation before evaluating the held-out gallery.

Figure~\ref{fig:per_value_recovery} shows the result for every value of shape, hue, and pattern.
Matched grounding improves average precision for 29 of the 30 values across all three factor vocabularies.
Although the gains vary in magnitude across values, their broadly positive pattern shows that matched image-side information can substantially improve linguistic access on held-out images.

\paragraph{Specificity of the visual contrast.}

\begin{wrapfigure}[11]{l}{0.43\textwidth}
    \centering
    \includegraphics[
        width=0.40\textwidth,
        trim={0.4cm 0.6cm 0.4cm 0.3cm}
    ]{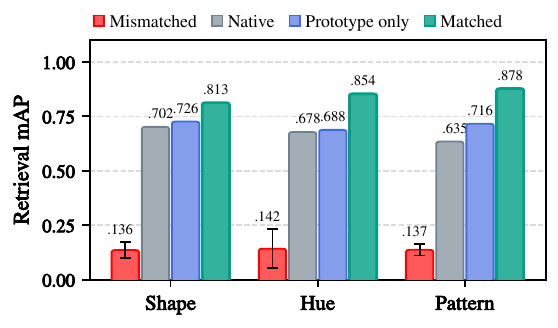}
    \caption{
    \textbf{Visual contrast matters.}
    Matched contrast outperforms prototype-only and mismatched-direction controls.
    }
    \label{fig:direction_specificity}
\end{wrapfigure}

The observed access gain raises a natural question: does it depend on the matched target-versus-rest visual contrast, or is moving the text query toward the target visual prototype sufficient?
We test the latter with \textit{prototype-only} grounding,
\[
q_v^{+}(\lambda)
=
\operatorname{norm}\left(
(1-\lambda)q_v+\lambda\mu_v^I
\right),
\]
with $\lambda$ selected on validation and fixed before test.

On SigLIP2 Base, prototype-only grounding yields modest gains over the native query but remains consistently below matched grounding (Fig.~\ref{fig:direction_specificity}). 
Moreover, assigning each target a nonmatching visual contrast sharply degrades retrieval, even when averaged over all nonmatching assignments. 
Together, these controls isolate what drives the access gain.
Rather than reflecting generic attraction toward target images, the improvement depends specifically on the matched visual direction that separates the target from its alternatives.
This interpretation is also consistent with the geometric guarantee, which applies to movement along the matched target-versus-rest visual contrast.

\paragraph{Prompt and lexical controls.}

\begin{wraptable}[10]{r}{0.45\textwidth}
    \vspace{-1.2em}
    \centering
    \scriptsize
    \setlength{\tabcolsep}{3pt}
    \caption{
    \textbf{Prompt robustness in the pattern setting.}
    Matched grounding improves across all tested query forms.
    }
    \label{tab:prompt_robustness_main}
    \begin{tabular}{p{0.30\textwidth}cc}
    \toprule
    Query form & Native & Matched \\
    \midrule
    \texttt{\{v\}} & .629 & .876 \\
    \texttt{a \{v\} surface} & .549 & .874 \\
    \texttt{an object with a \{v\} pattern} & .577 & .881 \\
    \texttt{a surface that is \{v\}} & .528 & .874 \\
    \texttt{a textured \{v\} surface} & .501 & .872 \\
    \bottomrule
    \end{tabular}
\end{wraptable}

The native query already averages three fixed prompt embeddings, reducing its dependence on any single wording.
We nevertheless test whether the observed access gain depends on this ensemble or on particular prompt formulations.
In the pattern setting, native mAP ranges from $0.501$ to $0.629$ across five fixed templates, whereas matched grounding improves performance for every template and remains between $0.872$ and $0.881$ (Table~\ref{tab:prompt_robustness_main}).
Thus, the access gain is not tied to the three-template ensemble or to any one tested formulation.

We also ask whether lexical substitution alone can close the access gap.
For each pattern, we form an equal-weight ensemble of the canonical term and two predetermined lexical alternatives.
This lexical ensemble reaches $0.512$ mAP, compared with $0.635$ for the original native ensemble, and therefore does not reproduce the improvement from matched grounding.
Together, these controls indicate that the observed access gains are not explained by prompt formulation or lexical alternatives.
The prompt-robustness result also holds for hue and shape, with full results reported in Appendix~\ref{app:prompt_controls}.

\paragraph{Dependence on visual evidence.}

\begin{wrapfigure}[15]{l}{0.40\textwidth}
    \centering
    \includegraphics[
        width=0.38\textwidth,
        trim={0.4cm 0.6cm 0.4cm 0.8cm}
    ]{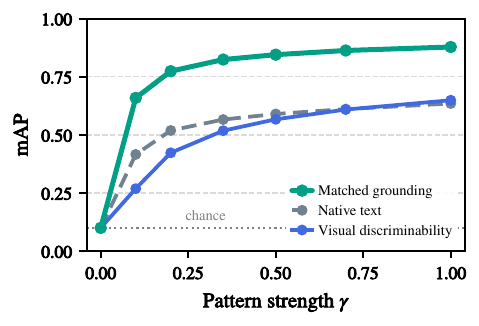}
    \caption{
    \textbf{Matched access gain under attenuated visual evidence.}
    Matched access gain remains substantial as the pattern signal weakens and disappears when the visual distinction is removed.
    }
    \label{fig:visual_evidence}
\end{wrapfigure}

We further extend the analysis by varying the strength of the underlying visual distinction for the pattern factor, where the visual signal can be directly controlled.
Each patterned image is progressively blended with its nuisance-matched plain counterpart, producing seven signal strengths from $\gamma=1.0$ to $\gamma=0.0$.
At each strength, visual directions are re-estimated from calibration images and the grounding strength $\alpha$ is selected on validation.

As shown in Fig.~\ref{fig:visual_evidence}, as the visual signal weakens, the available image-side contrast also decreases, and retrieval performance under matched grounding declines accordingly.
Even at $\gamma=0.1$, however, retrieval with matched-grounded queries reaches $0.659$ mAP, compared with $0.416$ for native-text retrieval, while the estimated visual directions retain a mean cosine of $0.825$ with their full-signal counterparts.
At $\gamma=0$, all pattern variants collapse to nuisance-matched plain images, and grounding provides no gain over chance.
These results indicate that matched access gains depend on the presence of the distinction-specific visual contrast itself.
Full results are provided in Appendix~\ref{app:visual_evidence_full}.

\subsection{Robustness Across Models and Held-Out Structure}
\label{sec:breadth}

\begin{table*}[h]
\centering
\scriptsize
\setlength{\tabcolsep}{5.8pt}
\renewcommand{\arraystretch}{1.1}
\caption{
\textbf{Matched access gain across factors, backbones, and held-out protocols.}
Each cell reports native $\rightarrow$ \textbf{matched} full-gallery mAP and factor-value top-1 accuracy.
Per-backbone rows use the context holdout; the lower block reports seven-backbone macro results across additional protocols.
}
\label{tab:breadth_full}

\begin{tabular}{l cc cc cc}
    \toprule
    & \multicolumn{2}{c}{Shape}
    & \multicolumn{2}{c}{Hue}
    & \multicolumn{2}{c}{Pattern} \\
    \cmidrule(lr){2-3}
    \cmidrule(lr){4-5}
    \cmidrule(lr){6-7}
    & mAP & Top-1
    & mAP & Top-1
    & mAP & Top-1 \\
    \midrule

    \rowcolor{gray!10}
    \multicolumn{7}{l}{\textit{Context holdout, per backbone}} \\

    \multicolumn{7}{l}{CLIP family} \\
    \quad CLIP ViT-B/16
    & $.519 \rightarrow \mathbf{.723}$ & $.543 \rightarrow \mathbf{.690}$
    & $.649 \rightarrow \mathbf{.809}$ & $.583 \rightarrow \mathbf{.836}$
    & $.361 \rightarrow \mathbf{.830}$ & $.425 \rightarrow \mathbf{.844}$ \\

    \quad EVA02-B/16
    & $.563 \rightarrow \mathbf{.804}$ & $.645 \rightarrow \mathbf{.767}$
    & $.586 \rightarrow \mathbf{.796}$ & $.656 \rightarrow \mathbf{.841}$
    & $.515 \rightarrow \mathbf{.843}$ & $.544 \rightarrow \mathbf{.884}$ \\

    \quad FG-CLIP Base
    & $.599 \rightarrow \mathbf{.740}$ & $.585 \rightarrow \mathbf{.706}$
    & $.678 \rightarrow \mathbf{.892}$ & $.720 \rightarrow \mathbf{.901}$
    & $.555 \rightarrow \mathbf{.848}$ & $.496 \rightarrow \mathbf{.882}$ \\

    \addlinespace[2pt]
    \multicolumn{7}{l}{SigLIP family} \\
    \quad SigLIP SO400M
    & $.730 \rightarrow \mathbf{.850}$ & $.741 \rightarrow \mathbf{.835}$
    & $.667 \rightarrow \mathbf{.825}$ & $.627 \rightarrow \mathbf{.856}$
    & $.663 \rightarrow \mathbf{.891}$ & $.718 \rightarrow \mathbf{.911}$ \\

    \quad SigLIP2 Base
    & $.702 \rightarrow \mathbf{.813}$ & $.692 \rightarrow \mathbf{.789}$
    & $.678 \rightarrow \mathbf{.854}$ & $.603 \rightarrow \mathbf{.882}$
    & $.635 \rightarrow \mathbf{.878}$ & $.676 \rightarrow \mathbf{.911}$ \\

    \quad SigLIP2 Large
    & $.744 \rightarrow \mathbf{.837}$ & $.770 \rightarrow \mathbf{.830}$
    & $.590 \rightarrow \mathbf{.736}$ & $.573 \rightarrow \mathbf{.756}$
    & $.620 \rightarrow \mathbf{.875}$ & $.653 \rightarrow \mathbf{.927}$ \\

    \quad SigLIP2 SO400M
    & $.764 \rightarrow \mathbf{.851}$ & $.774 \rightarrow \mathbf{.838}$
    & $.488 \rightarrow \mathbf{.592}$ & $.432 \rightarrow \mathbf{.620}$
    & $.621 \rightarrow \mathbf{.867}$ & $.676 \rightarrow \mathbf{.914}$ \\

    \midrule
    \rowcolor{gray!10}
    \multicolumn{7}{l}{\textit{Seven-backbone macro, per protocol}} \\

    Context holdout
    & $.660 \rightarrow \mathbf{.803}$ & $.678 \rightarrow \mathbf{.779}$
    & $.619 \rightarrow \mathbf{.786}$ & $.599 \rightarrow \mathbf{.813}$
    & $.567 \rightarrow \mathbf{.862}$ & $.598 \rightarrow \mathbf{.896}$ \\

    Unseen combinations
    & $.667 \rightarrow \mathbf{.814}$ & $.690 \rightarrow \mathbf{.798}$
    & $.618 \rightarrow \mathbf{.801}$ & $.592 \rightarrow \mathbf{.839}$
    & $.594 \rightarrow \mathbf{.876}$ & $.619 \rightarrow \mathbf{.906}$ \\

    24 nuisance partitions
    & $.663 \rightarrow \mathbf{.809}$ & $.687 \rightarrow \mathbf{.791}$
    & $.621 \rightarrow \mathbf{.788}$ & $.596 \rightarrow \mathbf{.818}$
    & $.570 \rightarrow \mathbf{.862}$ & $.603 \rightarrow \mathbf{.898}$ \\

    \bottomrule
\end{tabular}
\vspace{-1.2em}
\end{table*}


\paragraph{Across backbones and factors.}

We broaden the context-holdout audit to cover shape, hue, and pattern across seven frozen vision--language backbones.
Matched visual grounding improves both mAP and factor-value top-1 for all 21 backbone--factor combinations.
Though the gains vary in magnitude across models and factors, the improvement is consistent in every case.

\paragraph{Unseen semantic combinations.}

We also test a different kind of holdout based on semantic combinations.
Here, individual factor values remain present in calibration, validation, and test, but some combinations of the two non-target factors appear in only one split.
For example, when pattern is the target, evaluation uses hue--shape combinations that were not seen when the pattern directions were built or tuned.
By design, all nuisance contexts and renderer seeds appear in every split, so the evaluation isolates unseen semantic combinations rather than unseen nuisance conditions.
Matched visual grounding improves both metrics for all 21 backbone--factor combinations under this protocol.
The access gains therefore do not depend on memorizing particular semantic combinations and generalize to new combinations of otherwise familiar factor values.

\paragraph{Robustness to nuisance partitions.}

The context holdout used in our main analysis shows broad gains across backbone--factor combinations, but a single nuisance split could still be unusually favorable.
We therefore repeat the audit across 24 balanced partitions of the joint context--renderer realizations, assigning eight groups to calibration, four to validation, and twelve to test.
Visual directions are re-estimated separately for every partition.
Matched visual grounding improves both mAP and factor-value top-1 across all backbone--factor--partition configurations, with little variation across partitions.
This consistency indicates that the observed access gains are robust to how the available nuisance realizations are divided among calibration, validation, and test.

Taken together, these results show that matched access gains are not confined to a particular backbone--factor setting or held-out protocol.
Complete backbone-, protocol-, and value-resolved results are reported in Appendix~\ref{app:readouts}.

\subsection{Global Alignment and Residual Access Gaps}
\label{sec:global_alignment}

\vspace{-0.5em}
\begin{figure}[h]
    \centering
    \begin{minipage}[b]{0.53\linewidth}
        \centering
        \includegraphics[width=\linewidth]{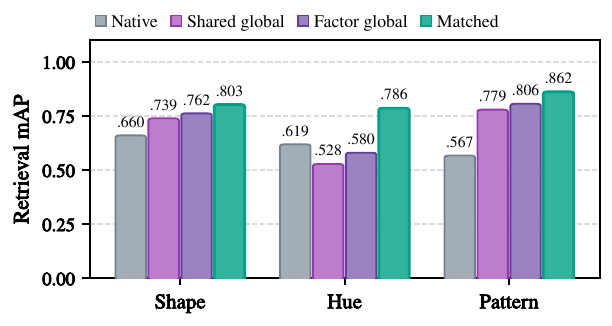}
    \end{minipage}
    \hfill
    \begin{minipage}[t]{0.43\linewidth}
        \centering
        \includegraphics[width=\linewidth]
        {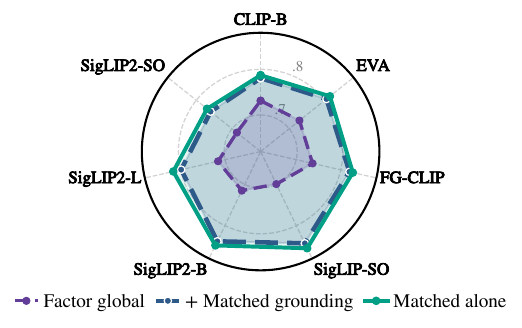}
    \end{minipage}
    \vspace{-0.08in}
    \caption{
    \textbf{Global alignment leaves residual access gaps.}
    (\textbf{Left}) Factor-level macro-mAP under the context holdout.
    (\textbf{Right}) Factor-global alignment before and after matched grounding across seven backbones, with native-space matched grounding shown for reference.
    }
    \label{fig:global_alignment}
    \vspace{-0.5em}
\end{figure}

We close this analysis by asking whether the access gaps identified for individual visual distinctions persist after a broader correction to the text--image representation geometry.
Such broad approaches are common in prior work on fine-grained cross-modal alignment, where correspondence is improved through shared transformations applied across many distinctions rather than distinction-specific corrections.
We test this using full-rank linear maps on the text representation, similar in spirit to prior work on linear cross-modal alignment~\citep{koishigarina2026clip,moayeri2023text}.

For a native query $q$, the aligned query is
\begin{equation}
q^{G}=\operatorname{norm}(Aq),
\end{equation}
where $A\in\mathbb{R}^{d\times d}$ is a bias-free, identity-initialized map and the pretrained encoders remain fixed.
We learn either a separate map $A_f$ for each factor or a single map $A_{\mathrm{shared}}$ across all three factors.
Both map types are trained on calibration image--caption pairs with factor-replacement hard negatives and selected on validation before test evaluation.
We use these maps as expressive baselines for broad cross-modal alignment.

Global alignment improves overall retrieval but leaves substantial residual access gaps across factors.
Under the context holdout, seven-backbone macro-mAP rises from $0.616$ natively to $0.682$ with shared-global alignment and $0.716$ with factor-global alignment.
Both maps improve pattern and shape mAP but reduce hue mAP, whereas matched visual grounding achieves the highest mAP on all three factors (Fig.~\ref{fig:global_alignment}, left).

We next test whether the value-specific visual structure remains useful even after broad alignment.
To examine this, we fix each global map and use matched visual directions estimated from calibration images to ground the aligned queries, with the grounding strength reselected on validation.
Factor-global alignment provides the stronger broad correction, yet matched visual grounding further raises macro-mAP from $0.716$ to $0.806$ and yields additional access gains across all seven backbones (Fig.~\ref{fig:global_alignment}, right).
The same overall trend holds across additional global alignment baselines and broader held-out evaluations (Appendix~\ref{app:global_alignment_full}).
Thus, even after a broad correction to text--image alignment, matched grounding that targets the value-specific visual contrast can further reduce the remaining access gaps.


\section{Consequences and Generalization of Improved Access}
\label{sec:consequences}

Having established gaps between visual discriminability and linguistic addressability, together with consistent gains from matched visual grounding, we next turn to the broader consequences of improved access and its generalization beyond the controlled setting.

\subsection{From Factor Access to Compositional Retrieval}
\label{sec:composition}

\begin{figure}[h]
    \centering
    \includegraphics[
        width=\linewidth,
        trim={0.2cm 0.2cm 0.2cm 0.2cm},
        clip
    ]{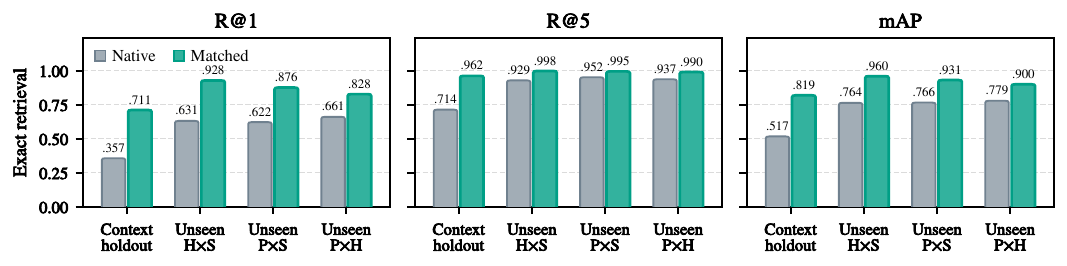}
    \caption{
    \textbf{Factor-level access gains support exact compositional retrieval.}
    Native and matched queries are compared under the context holdout and three unseen-combination protocols.
    Matched visual grounding improves exact R@1, R@5, and mAP across all held-out conditions.
    H$\times$S, P$\times$S, and P$\times$H denote held-out hue--shape, pattern--shape, and pattern--hue combinations.
    }
    \label{fig:exact_composition}
\end{figure}

Matched grounding is local to the visual contrast associated with each factor value, so improved access to an individual factor value does not by itself establish that the resulting queries remain compatible when several distinctions must be satisfied jointly.
We therefore evaluate exact composition, where a query specifies one pattern, one hue, and one shape and the correct image must match all three values simultaneously.
The exact target must outrank near-miss images that differ in one or more factors.

Following prior work showing the utility of factorized inference for compositional retrieval~\citep{alshehri2026similarity}, we score each requested factor value within its own vocabulary before combining the resulting factor-level evidence.
Native and matched queries use the same composition rule, so the comparison isolates the effect of improved factor-level access rather than differences in the composition operator.
Specifically, for each factor $f$, $p_f(v\mid x)$ denotes the softmax-normalized similarity of image $x$ to value $v$ within that factor vocabulary, using a shared temperature $\tau$.

For each factor $f$, we define
\[
p_f(v\mid x)
=
\frac{
\exp\!\left(\langle q_{f,v},x\rangle/\tau\right)
}{
\sum_{u\in V_f}
\exp\!\left(\langle q_{f,u},x\rangle/\tau\right)
},
\qquad
S(y,x)=
\sum_{f\in\{\mathrm{shape},\mathrm{hue},\mathrm{pattern}\}}
w_f\log p_f(y_f\mid x).
\]

Each factor-specific grounding strength $\alpha_f$ is chosen on factor-level validation and then fixed.
The shared temperature $\tau$ and factor weights $w_f$ are chosen on compositional validation to balance how factor-level evidence is combined before test evaluation.

In this setting, matched visual grounding substantially improves exact retrieval across all held-out conditions and metrics (Fig.~\ref{fig:exact_composition}).
Under the context holdout, R@1 rises from $0.357$ to $0.711$, R@5 from $0.714$ to $0.962$, and mAP from $0.517$ to $0.819$.
The gains also persist under unseen-combination protocols, where individual factor values remain familiar but their test-time combinations are new.
The factor-level access gains therefore carry over to exact compositional retrieval, enabling new multi-factor combinations to be retrieved jointly rather than only improving isolated factor readouts.

\subsection{Validation-Guided Selective Grounding}
\label{sec:selective}

\begin{wraptable}[11]{r}{0.30\textwidth}
    \vspace{-1.0em}
    \centering
    \small
    \setlength{\tabcolsep}{3.5pt}
    \caption{
    \textbf{Validation-guided selective grounding.}
    Test mAP across the context and unseen-combination protocols.
    }
    \label{tab:selective_grounding}
    \vspace{-0.5em}
    \begin{tabular}{lc}
        \toprule
        Policy & Test mAP \\
        \midrule
        Native & .6797 \\
        Ground every value & .8220 \\
        Validation-guided & \textbf{.8245} \\
        \bottomrule
    \end{tabular}
\end{wraptable}

Because access gains vary across individual factor values, the value-level audit can also guide where grounding should be applied.
Some native queries are already effective, while others show clear validation gains from matched grounding.
We therefore retain grounding only for values whose validation AP gain remains positive under bootstrap resampling of nuisance groups; otherwise, we keep the native query.
A single grounding strength is selected per factor on validation, and all decisions are fixed before test evaluation.

On SigLIP2 Base, across the context-holdout and unseen hue--shape-combination protocols, grounding every value raises mean test mAP from $0.6797$ to $0.8220$, while the validation-guided policy reaches $0.8245$. 
Treating each value-specific visual distinction separately enables more targeted grounding, applying it only where validation supports a gain while yielding a small additional improvement.

\subsection{Matched Access Gains Beyond FactorAtlas}
\label{sec:natural}

\begin{table}[h]
\centering
\small
\setlength{\tabcolsep}{4pt}
\renewcommand{\arraystretch}{1.08}
\caption{
\textbf{Natural-image transfer settings.}
We evaluate texture, garment pattern and length, and material distinctions using dataset-specific splits that separate evaluation examples from those used to estimate and tune the grounding.
}
\label{tab:natural_settings}
\vspace{-0.8em}
\begin{tabular*}{\linewidth}{
    @{\extracolsep{\fill}}
    lll
    @{}
}
\toprule
Dataset & Target distinction & Held-out setting \\
\midrule
DTD~\citep{cimpoi2014describing}
& Texture
& Official partitions \\

Fashionpedia~\citep{jia2020fashionpedia}
& Pattern / garment length
& Category-held-out / image-disjoint \\

COCO-Facet~\citep{li2026highlighting}
& Material
& Image-disjoint \\

UT-Zappos~\citep{fine-grained}
& Material
& Product-disjoint \\
\bottomrule
\end{tabular*}
\vspace{-0.8em}
\end{table}

\begin{wrapfigure}[16]{l}{0.50\textwidth}
    \centering
    \includegraphics[
        width=0.49\textwidth,
        trim={0.2cm 0.2cm 0.2cm 0.2cm},
        clip
    ]{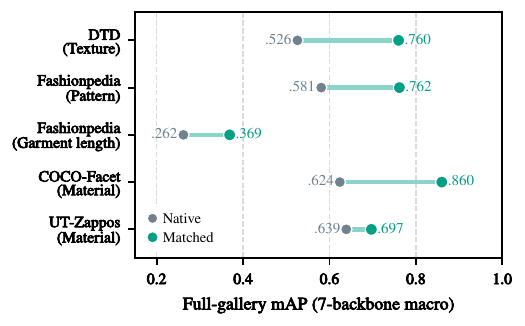}
    \caption{
    \textbf{Matched Access Gain beyond \mbox{FactorAtlas}.}
    Seven-backbone macro mAP improves across natural texture, garment, and material distinctions.
    }
    \label{fig:natural_transfer}
\end{wrapfigure}

We finally evaluate matched visual grounding beyond the controlled FactorAtlas setting using natural images.
To do so, we draw on datasets that support fine-grained vocabularies for visual factors such as texture, garment pattern and length, and material, and construct controlled text-to-image retrieval settings with dataset-specific splits (Table~\ref{tab:natural_settings}).
Despite substantial within-dataset variation in how target values are visually realized, matched grounding yields consistent access gains across the evaluated settings, with mAP improving from $0.262$ to $0.369$ for Fashionpedia garment length and from $0.624$ to $0.860$ for COCO-Facet material when averaged across seven backbones (Fig.~\ref{fig:natural_transfer}).
These results show that the diagnosis and intervention developed in the controlled setting also carry over to natural images.

We further examine whether visual contrasts derived in FactorAtlas can themselves transfer to natural images.
For the four pattern values shared with Fashionpedia, directly applying the corresponding contrasts and grounding strengths raises mAP from $0.583$ to $0.719$ across seven backbones, without target-domain fitting or validation.
This suggests that visual structure derived for a well-defined distinction in a controlled setting can remain useful when the same distinction recurs in natural images.
Full results and protocols are reported in Appendix~\ref{app:natural_full}.

\section{Relation to Prior Work}
\label{sec:related}

\paragraph{Fine-grained and compositional vision--language retrieval.}
Vision--language models such as CLIP and SigLIP enable retrieval and zero-shot recognition through cross-modal similarity~\citep{radford2021learning,zhai2023sigmoid}.
Yet benchmarks such as Winoground, VL-CheckList, SugarCrepe, and COLA reveal limitations of native vision--language matching on fine-grained attributes, relations, binding, and composition~\citep{thrush2022winoground,zhao2022vl,hsieh2023sugarcrepe,ray2023cola}.
Recent work has further examined these challenges in multi-condition retrieval, attribute binding, and factorized inference~\citep{chow2026merit,lu2026beyond,zhang2025abe,alshehri2026similarity}.
Complementing these lines of work, we ask whether weak retrieval for a named visual distinction reflects limited image-side discriminability or limited access through the native text query.

\paragraph{Representation, hidden structure, and linguistic access.}
Shared VLM embedding spaces exhibit meaningful semantic structure across image and text representations~\citep{papadimitriou2025interpreting}.
On the image side in particular, frozen embeddings preserve compositional organization and ordinal directions identifiable from visual examples~\citep{berasi2025not,sonthalia2026rankability}.
Yet information preserved within a modality may remain weakly accessible through native image--text similarity when image and text representations are not well aligned~\citep{koishigarina2026clip}.
Building on this gap, we take each named visual distinction as the unit of analysis, enabling a finer-grained audit of the relation between image-side structure and linguistic access.

\paragraph{Cross-modal alignment and distinction-specific access.}
Prior work studies broad cross-modal mismatch~\citep{liang2022mind} and approaches for improving image--text correspondence through learned mappings, structure-preserving alignment, or stronger modality-specific representations~\citep{moayeri2023text,eslami2024mitigate,groger2026limited,gong2025kernel,huang2026llm2clip}.
Among these approaches, LABCLIP improves attribute--object correspondence through a learned transformation shared across text embeddings~\citep{koishigarina2026clip}.
Support-based adaptation provides another related direction, using labeled visual examples to improve downstream performance with frozen vision--language models~\citep{zhang2021tip}.
In our approach, we focus on distinction-specific access, deriving a matched target-versus-rest contrast for each visual value and using it to adjust the native text query on held-out data.
This value-specific intervention tests whether the corresponding image-side structure can reduce the access gap, including after global alignment.


\section{Conclusion}
\label{sec:conclusion}

A visual distinction can remain discriminable in frozen image geometry while being only weakly addressed by its native text query, so retrieval failure need not imply that the corresponding distinction is absent from the visual representation. Matched visual grounding provides a held-out intervention: when the relevant image-side contrast is present, using that contrast can improve linguistic access and reduce the observed access gap.

This interpretation is supported by direction-specific and visual-evidence controls, robustness across models and held-out conditions, and residual gains after global alignment.
Improved access to individual factor values supports exact compositional retrieval and validation-guided selective grounding, while the same matched-grounding logic extends to natural-image distinctions.
More broadly, strong aggregate cross-modal alignment does not by itself ensure reliable linguistic access to every distinction that remains discriminable in image geometry. 

These conclusions are conditioned on visual discriminability.
Access gains vary across settings, the geometric guarantee is local, and the global maps cover only one family of broader alignment corrections.
Within this scope, what frozen image geometry can reliably distinguish and what the native language interface can reliably address should be treated as separate empirical questions.
When a distinction remains visually discriminable but linguistically under-addressed, matched visual grounding provides a held-out intervention that tests this diagnosis and can improve linguistic access.
Extending this perspective beyond predefined distinctions and labeled visual support could enable vision--language systems to more flexibly expose and use visual structure that their representations already support, even when that structure is not well captured by the native language interface.


\bibliography{iclr2027_conference}
\bibliographystyle{iclr2027_conference}

\clearpage
\appendix

\section{Full Proof of the Geometric Guarantee}
\label{app:proof}

Let $\|q\|=\|d\|=1$, $\Delta=\|\Delta\|d$, and define
\[
q_\alpha=\frac{q+\alpha d}{\|q+\alpha d\|}
\]
whenever $q+\alpha d\neq0$.
For any unit query $r$, define the matched score gap
\[
M(r)=r^\top\Delta.
\]

\paragraph{Proposition.}
For every $\alpha\geq0$ for which $q_\alpha$ is defined,
\[
M(q_\alpha)\geq M(q).
\]

\paragraph{Proof.}
Let $c=q^\top d\in[-1,1]$. Since
\[
\|q+\alpha d\|^2=1+2\alpha c+\alpha^2,
\]
we obtain
\[
M(q_\alpha)
=
\|\Delta\|
\frac{c+\alpha}{\sqrt{1+2\alpha c+\alpha^2}}.
\]

For $|c|<1$,
\[
\frac{d}{d\alpha}M(q_\alpha)
=
\|\Delta\|
\frac{1-c^2}{(1+2\alpha c+\alpha^2)^{3/2}}
>0,
\]
so the matched score gap increases strictly with $\alpha$.

For $c=1$, we have $q=d$, hence $q_\alpha=d$ for every $\alpha\geq0$, and therefore
\[
M(q_\alpha)=M(q)=\|\Delta\|.
\]

For $c=-1$, $q=-d$ and
\[
q_\alpha=
\begin{cases}
-d, & 0\leq\alpha<1,\\
\text{undefined}, & \alpha=1,\\
d, & \alpha>1,
\end{cases}
\]
so
\[
M(q_\alpha)=
\begin{cases}
-\|\Delta\|, & 0\leq\alpha<1,\\
+\|\Delta\|, & \alpha>1.
\end{cases}
\]
Thus $M(q_\alpha)\geq M(q)$ for every defined $\alpha\geq0$.
\hfill$\square$

\paragraph{Negative initial gap.}
If $-1<c<0$, then $M(q)<0$, and the grounded score gap crosses zero at
\[
\alpha^\star=-c=-q^\top d.
\]
Hence
\[
M(q_{\alpha^\star})=0,
\qquad
M(q_\alpha)>0
\quad\text{for }\alpha>\alpha^\star.
\]
The singular endpoint $c=-1$ is excluded from this continuous crossing statement.

\paragraph{Scope.}
The guarantee concerns only the matched calibration target-versus-average-rest gap.
It does not imply monotonic improvement in held-out retrieval, preservation of unrelated factors, or successful composition; these are evaluated empirically.

\section{FactorAtlas Construction and Evaluation Protocol}
\label{app:factoratlas_protocol}

FactorAtlas is the complete product of 10 patterns, 12 hues, 8 shapes, 12 nuisance contexts, and two renderer replicas, totaling 23,040 images.
Contexts 0--3 are calibration (7,680 images), 4--5 validation (3,840), and 6--11 test (11,520); both replicas of a context remain in the same split.
Contexts vary pose, viewpoint, illumination, background, roughness, and material appearance. Images are rendered at $224\times224$ using Cycles. 
Camera elevation is sampled in $[0.14,0.98]$ radians, camera distance in $[3.65,6.35]$, lens in $[42,66]$, object pitch and roll in $[-0.72,0.72]$, yaw in $[0,2\pi]$, world strength in $[0.08,0.72]$, and material roughness in $[0.25,0.90]$. 
Each scene uses three randomized area lights.
Calibration constructs prototypes and visual contrasts; validation selects all hyperparameters; test is evaluated after those choices are fixed.

The native query is the normalized mean of three prompts:
\texttt{\{v\}}, \texttt{a \{v\} surface}, and
\texttt{an object with a \{v\} pattern} for pattern;
\texttt{\{v\}}, \texttt{a \{v\} object}, and
\texttt{a \{v\} surface} for hue; and
\texttt{\{v\}}, \texttt{a \{v\}}, and
\texttt{an object shaped like a \{v\}} for shape.
$\alpha\in\{0,.05,\ldots,8\}$ is selected by validation macro-mAP, breaking ties by factor-value top-1 and then smaller $\alpha$. 
Macro-mAP averages per-value full-gallery AP; factor-value top-1 is an image-wise $K$-way readout and is distinct from retrieval R@1.

All encoders are frozen. We use official preprocessing for
CLIP ViT-B/16~\citep{radford2021learning},
EVA02-B/16~\citep{sun2023eva},
FG-CLIP Base~\citep{xie2025fg},
SigLIP SO400M~\citep{zhai2023sigmoid},
SigLIP2 Base, SigLIP2 Large, and SigLIP2 SO400M~\citep{tschannen2025siglip}.
Exact checkpoints are respectively \texttt{openai/clip-vit-base-patch16}, OpenCLIP \texttt{EVA02-B-16} with \texttt{merged2b\_s8b\_b131k}, \texttt{qihoo360/fg-clip-base}, \texttt{google/siglip-so400m-patch14-384}, \texttt{google/siglip2-base-patch16-224}, \texttt{google/siglip2-large-patch16-256}, and \texttt{google/siglip2-so400m-patch16-384}.

\section{Full Diagnostic Controls}
\label{app:diagnostic_controls}

\subsection{Intervention-Free Separation}

We begin with the comparison that does not modify the text query. 
For each factor value, the image-prototype query is the normalized centroid estimated only from labeled calibration images, whereas the native query is produced by the frozen text encoder from the prompt ensemble specified in Appendix~\ref{app:factoratlas_protocol}. 
Both queries retrieve the same held-out test gallery. 
The comparison therefore asks whether a distinction remains discriminable in calibration-derived image geometry even when the frozen text interface addresses it less reliably; it is not a comparison between two trained classifiers.

\begin{table}[h]
\centering\small
\caption{\textbf{Intervention-free factor-value top-1.}}
\label{tab:factor_separation_full}
\begin{tabular}{lccc}\toprule
 & Pattern & Hue & Shape\\\midrule
\multicolumn{4}{l}{\textit{SigLIP2 Base}}\\
Image prototype&.919&.909&.825\\ Native text&.676&.603&.692\\
\multicolumn{4}{l}{\textit{Seven-backbone macro}}\\
Image prototype&.907&.835&.815\\ Native text&.598&.599&.678\\\bottomrule
\end{tabular}
\end{table}

Table~\ref{tab:factor_separation_full} reports the image-wise factor-value top-1 readout. 
The image prototype exceeds the native text query for pattern, hue, and shape on SigLIP2 Base, and the same ordering remains after averaging the seven backbones. 
The gap is largest for pattern in the seven-backbone macro ($.907$ versus $.598$), but is also present for hue and shape. 
This is the intervention-free evidence that visual discriminability can exceed native linguistic addressability; matched grounding has not yet been applied in this table.

\subsection{Specificity of the Visual Distinction}
\label{app:visual_distinction}

The main results raise a stronger question: is the target-versus-rest visual distinction itself important, or is it sufficient to pull the text query toward images of the correct value?
We test the latter possibility with target-prototype attraction.
For each value $v$, we interpolate the native text query toward its calibration-derived target image centroid:
\[
q_v^{+}(\lambda)
=
\operatorname{norm}\left(
(1-\lambda)q_v+\lambda\mu_v^I
\right).
\]
This baseline uses the correct target-image support but does not contrast the target value against the remaining values in its factor vocabulary.
By comparison, target-versus-rest grounding uses the direction
\[
d_v^I
=
\operatorname{norm}\left(
\mu_v^I-\bar{\mu}_{-v}^I
\right),
\]
which explicitly represents the visual distinction between the target value and its alternatives.

For each backbone and factor, a single $\lambda$ shared by all values is selected from $\{0,0.0125,\ldots,1\}$ using validation mAP and then fixed before test evaluation.
The grounding strength $\alpha$ is selected independently on the same validation split.
Both methods use the same calibration support, native queries, held-out gallery, and evaluation metric.

\begin{table*}[h]
\centering
\scriptsize
\setlength{\tabcolsep}{4pt}
\renewcommand{\arraystretch}{1.08}
\caption{
\textbf{Target attraction versus target-versus-rest visual contrast.}
Each factor cell reports native $\rightarrow$ target-prototype attraction $\rightarrow$ \textbf{target-versus-rest grounding} mAP under the primary context holdout.
The final column reports the validation-selected attraction strengths for pattern, hue, and shape.
}
\label{tab:prototype_attraction_full}
\begin{tabular}{lcccc}
\toprule
Backbone
& Pattern
& Hue
& Shape
& $\lambda_{\mathrm{P/H/S}}$ \\
\midrule
CLIP ViT-B/16
& $.361 \rightarrow .568 \rightarrow \mathbf{.830}$
& $.649 \rightarrow .656 \rightarrow \mathbf{.809}$
& $.519 \rightarrow .584 \rightarrow \mathbf{.723}$
& $.713/.163/.500$ \\

EVA02-B/16
& $.515 \rightarrow .643 \rightarrow \mathbf{.843}$
& $.586 \rightarrow .595 \rightarrow \mathbf{.796}$
& $.563 \rightarrow .649 \rightarrow \mathbf{.804}$
& $.525/.150/.488$ \\

FG-CLIP Base
& $.555 \rightarrow .689 \rightarrow \mathbf{.848}$
& $.678 \rightarrow .708 \rightarrow \mathbf{.892}$
& $.599 \rightarrow .653 \rightarrow \mathbf{.740}$
& $.700/.363/.500$ \\

SigLIP SO400M
& $.663 \rightarrow .745 \rightarrow \mathbf{.891}$
& $.667 \rightarrow .683 \rightarrow \mathbf{.825}$
& $.730 \rightarrow .765 \rightarrow \mathbf{.850}$
& $.613/.175/.538$ \\

SigLIP2 Base
& $.635 \rightarrow .716 \rightarrow \mathbf{.878}$
& $.678 \rightarrow .688 \rightarrow \mathbf{.854}$
& $.702 \rightarrow .726 \rightarrow \mathbf{.813}$
& $.425/.100/.400$ \\

SigLIP2 Large
& $.620 \rightarrow .732 \rightarrow \mathbf{.875}$
& $.590 \rightarrow .597 \rightarrow \mathbf{.736}$
& $.744 \rightarrow .756 \rightarrow \mathbf{.837}$
& $.450/.113/.213$ \\

SigLIP2 SO400M
& $.621 \rightarrow .726 \rightarrow \mathbf{.867}$
& $.488 \rightarrow .497 \rightarrow \mathbf{.592}$
& $.764 \rightarrow .772 \rightarrow \mathbf{.851}$
& $.463/.100/.138$ \\
\midrule
Seven-backbone macro
& $.567 \rightarrow .689 \rightarrow \mathbf{.862}$
& $.619 \rightarrow .632 \rightarrow \mathbf{.786}$
& $.660 \rightarrow .701 \rightarrow \mathbf{.803}$
& -- \\
\bottomrule
\end{tabular}
\end{table*}

Target-prototype attraction improves over native retrieval in all 21 backbone--factor cases, showing that target attraction alone can yield gains.
However, target-versus-rest grounding performs better than prototype attraction in all 21 cases.
Across factors and backbones, macro-mAP increases from $.616$ under native retrieval to $.674$ with prototype attraction and $.817$ with target-versus-rest grounding.
Thus, target attraction alone does not explain the observed access gain; explicitly isolating the visual distinction between each value and its alternatives yields substantially larger held-out gains.
This held-out result is separate from, but consistent with, the local geometric guarantee for movement along the matched target-versus-average-rest calibration gap.

\subsection{Prompt and Lexical Controls}
\label{app:prompt_controls}

We test whether matched access gain can be explained by the wording of the native text query.
The detailed tables below use the frozen SigLIP2 Base backbone and the primary context holdout.
Visual directions are estimated from calibration contexts 0--3, grounding strength is selected on validation contexts 4--5, and all results are reported on test contexts 6--11.

\subsubsection{Fixed Prompt Templates}

For each factor, we define five grammatical templates that express the same factor value in different ways.
Every template is applied uniformly to the complete factor vocabulary; no template is selected separately for individual values.
Table~\ref{tab:prompt_template_bank} lists the complete template bank.

\begin{table*}[h]
\centering
\scriptsize
\setlength{\tabcolsep}{5pt}
\caption{
\textbf{Fixed factor-specific prompt templates.}
The placeholder \texttt{\{v\}} is replaced by every value in the corresponding factor vocabulary.
}
\label{tab:prompt_template_bank}
\begin{tabular}{clll}
\toprule
ID & Pattern & Hue & Shape \\
\midrule
T1
& \texttt{\{v\}}
& \texttt{\{v\}}
& \texttt{\{v\}} \\

T2
& \texttt{a \{v\} surface}
& \texttt{a \{v\} object}
& \texttt{a \{v\}} \\

T3
& \texttt{an object with a \{v\} pattern}
& \texttt{an object that is \{v\}}
& \texttt{an object shaped like a \{v\}} \\

T4
& \texttt{a surface that is \{v\}}
& \texttt{a \{v\} colored object}
& \texttt{a geometric \{v\}} \\

T5
& \texttt{a textured \{v\} surface}
& \texttt{a \{v\} surface}
& \texttt{a three dimensional \{v\} object} \\
\bottomrule
\end{tabular}
\end{table*}

For each factor--template combination, we compare the native query with its matched-grounded counterpart.
A single factor-level $\alpha$ is selected on validation separately for each template and fixed before test evaluation.
Table~\ref{tab:prompt_template_results} reports all 15 test endpoints.
Matched grounding improves over the corresponding native query in every case.

\begin{table}[h]
\centering
\small
\setlength{\tabcolsep}{5.5pt}
\caption{
\textbf{Prompt robustness on SigLIP2 Base.}
Each entry reports native $\rightarrow$ matched test mAP.
}
\label{tab:prompt_template_results}
\begin{tabular}{cccc}
\toprule
Template & Pattern & Hue & Shape \\
\midrule
T1 & $.629 \rightarrow .876$ & $.657 \rightarrow .846$ & $.704 \rightarrow .813$ \\
T2 & $.549 \rightarrow .874$ & $.636 \rightarrow .853$ & $.696 \rightarrow .813$ \\
T3 & $.577 \rightarrow .881$ & $.646 \rightarrow .853$ & $.685 \rightarrow .813$ \\
T4 & $.528 \rightarrow .874$ & $.656 \rightarrow .862$ & $.642 \rightarrow .813$ \\
T5 & $.501 \rightarrow .872$ & $.585 \rightarrow .846$ & $.659 \rightarrow .813$ \\
\midrule
Improved & $5/5$ & $5/5$ & $5/5$ \\
\bottomrule
\end{tabular}
\end{table}

The validation-selected strengths for T1--T5 are $[.95,1.00,.90,1.05,1.40]$ for pattern, $[.80,.80,.85,.65,1.05]$ for hue, and $[2.40,2.70,3.00,2.95,3.55]$ for shape.
The result is therefore not tied to any of the tested query forms.
Across all seven audited backbones, matched grounding improves all $7\times3\times5=105$ backbone--factor--template endpoints under the same per-template evaluation.

\subsubsection{Pattern Lexical Substitution}

We separately test whether lexical substitution alone can close the observed pattern access gap.
For each pattern, we construct an equal-weight ensemble from three predetermined lexical forms using the template \texttt{a TERM surface}.
No term is selected using validation or test performance.
This experiment does not apply matched grounding to the lexical variants; it tests whether replacing the words alone can reproduce the observed access gain.

\begin{table*}[h]
\centering
\small
\setlength{\tabcolsep}{5pt}
\caption{
\textbf{Pattern lexical-substitution control.}
The canonical native ensemble is compared with an equal-weight ensemble of three fixed lexical forms.
}
\label{tab:lexical_control}
\begin{tabular}{lllcc}
\toprule
Value & Additional lexical forms & & Canonical AP & Lexical AP \\
\midrule
Plain      & solid, unpatterned            && .287 & .109 \\
Striped    & banded, lined                 && .375 & .367 \\
Wavy       & undulating, wave patterned    && .666 & .558 \\
Checkered  & checkerboard, chequered       && .999 & .976 \\
Dotted     & polka dotted, dot patterned   && .711 & .490 \\
Cracked    & fissured, crackle patterned   && .936 & .744 \\
Mottled    & blotchy, variegated           && .669 & .345 \\
Speckled   & flecked, peppered             && .071 & .084 \\
Marbled    & veined, marble patterned      && .815 & .687 \\
Brick      & brickwork, brick patterned    && .815 & .758 \\
\midrule
\multicolumn{3}{l}{Macro mAP}              & .635 & .512 \\
\multicolumn{3}{l}{Label top-1}            & .676 & .495 \\
\bottomrule
\end{tabular}
\end{table*}

Lexical substitution decreases aggregate mAP from $0.635$ to $0.512$ and decreases AP for nine of the ten pattern values.
Together, the two controls show that matched access gain persists across the tested factor-specific prompt formulations, while lexical substitution alone does not reproduce it.

\subsection{Visual-Evidence Dependence}
\label{app:visual_evidence_full}

The preceding controls establish direction specificity, but labeled calibration support could still appear to produce gains independently of a repeatable visual contrast. 
To test dependence on that signal, each patterned image is blended with a nuisance-matched plain rendering of the same scene.
Thus pose, viewpoint, illumination, shape, hue, and background are held fixed while the pattern evidence is progressively attenuated. 
A strength of $\gamma=1$ is the original patterned image, and $\gamma=0$ is the matched plain image. 
Visual directions are reconstructed from calibration and $\alpha$ is reselected on validation independently at every strength, so intermediate points do not reuse the full-signal direction.

\begin{table}[h]
\centering\small
\caption{\textbf{Pattern attenuation on SigLIP2 Base.} 
Performance is mAP; cosine is measured against the full-signal directions.}
\label{tab:attenuation_full}
\begin{tabular}{ccccc}\toprule
$\gamma$&Image query&Native&Matched&Cosine\\\midrule
1.00&.649&.635&.878&1.000\\ .70&.609&.611&.863&.995\\
.50&.567&.590&.845&.981\\ .35&.518&.566&.824&.962\\
.20&.423&.519&.774&.920\\ .10&.269&.416&.659&.825\\
.00&.100&.100&.100&-.001\\\bottomrule
\end{tabular}
\end{table}

As the signal weakens, image-side discriminability and the absolute level of retrieval achieved by matched grounding decline rather than failing abruptly.
At $\gamma=.10$, the image-side direction is weaker but remains aligned with its full-signal counterpart (mean cosine $.825$), and matched mAP remains $.659$ compared with native $.416$. 
At $\gamma=0$, images with different pattern labels are identical within each nuisance-matched group, all three readouts reach the ten-way chance value of $.100$, and the matched access gain disappears. 
The three readouts use different queries and therefore are not upper bounds on one another; in particular, matched mAP may exceed image-prototype mAP. 
The relevant result is their shared dependence on a repeatable visual contrast: weak but stable evidence can support matched access gains, whereas labels alone cannot do so after that contrast is removed.

\subsection{Finite Grounding and the Pure Visual Endpoint}
\label{app:pure_visual_endpoint}

Our main analysis treats matched visual grounding as an intervention on the native text query.
The native query $q_v$ is moved along the value-specific visual contrast $d_v^I$, with a factor-level grounding strength selected on validation.
A possible limiting interpretation is that the resulting gains arise simply because the grounded query approaches an image-derived visual readout, effectively discarding the native text query.

We therefore characterize the relation between finite matched grounding and this limiting endpoint.
This is not a separate adaptation method or an additional leaderboard comparison.
Rather, it is an endpoint analysis of the same intervention family defined in Eq.~(1).

\paragraph{The intervention path.}

Recall that matched grounding is defined as
\begin{equation}
q_v'(\alpha)
=
\operatorname{norm}\!\left(q_v+\alpha d_v^I\right),
\qquad
\alpha \ge 0.
\label{eq:app_grounding_path}
\end{equation}

The two endpoints are
\begin{equation}
q_v'(0)=q_v
\end{equation}
and
\begin{equation}
\lim_{\alpha\rightarrow\infty} q_v'(\alpha)=d_v^I.
\end{equation}

Thus, the native text query, the validation-selected finite grounded query, and the pure visual direction lie on the same intervention path:
\begin{equation}
\underbrace{q_v}_{\alpha=0\text{: native text}}
\quad\longrightarrow\quad
\underbrace{q_v'(\widehat{\alpha})}_{\text{validation-selected finite grounding}}
\quad\longrightarrow\quad
\underbrace{d_v^I}_{\alpha\rightarrow\infty\text{: pure visual endpoint}}.
\end{equation}

The pure visual direction $d_v^I$ is therefore the limiting endpoint of the same query family rather than an unrelated baseline.

\paragraph{Geometric location of a finite grounded query.}

The grounding strength also has a simple geometric interpretation.
Let $q$ and $d$ be unit vectors with $q\neq -d$, and let $\alpha\ge 0$.
Define
\begin{equation}
q_\alpha
=
\frac{q+\alpha d}{\|q+\alpha d\|}
\end{equation}
and let
\begin{equation}
c=q^\top d.
\end{equation}

Then
\begin{equation}
q_\alpha^\top q
=
\frac{1+\alpha c}
{\sqrt{1+\alpha^2+2\alpha c}},
\end{equation}
while
\begin{equation}
q_\alpha^\top d
=
\frac{c+\alpha}
{\sqrt{1+\alpha^2+2\alpha c}}.
\end{equation}

Subtracting gives
\begin{equation}
q_\alpha^\top q-q_\alpha^\top d
=
\frac{(1-\alpha)(1-c)}
{\sqrt{1+\alpha^2+2\alpha c}}.
\label{eq:app_native_visual_distance}
\end{equation}

\paragraph{Proposition.}

Let $q$ and $d$ be distinct, non-antipodal unit vectors and let
\begin{equation}
q_\alpha=\operatorname{norm}(q+\alpha d).
\end{equation}

Then $q_\alpha$ is closer in cosine similarity to $q$ than to $d$ whenever $0\le\alpha<1$.
It is equally close to the two directions at $\alpha=1$.
It is closer in cosine similarity to $d$ than to $q$ whenever $\alpha>1$.

\paragraph{Proof.}

Because $q\neq d$, we have $c=q^\top d<1$.
Therefore, $1-c>0$.
The denominator in Eq.~\ref{eq:app_native_visual_distance} is positive whenever $q_\alpha$ is defined.
Hence, the sign of $q_\alpha^\top q-q_\alpha^\top d$ is determined entirely by $1-\alpha$.
The three cases follow immediately.
\hfill$\square$

This proposition describes only the geometric position of the grounded query along the intervention path.
It does not assign a fraction of retrieval performance to the text or visual component.
Accordingly, $\alpha$ should not be interpreted as an attribution weight.

\paragraph{Empirical behavior under limited calibration support.}

We compare finite grounding with the corresponding pure visual endpoint as the amount of calibration support used to estimate the visual contrast varies.
For each factor value, we subsample
$n\in\{1,2,4,8,16,32,64,128,256\}$
calibration images.
Each budget uses 12 balanced resamples; visual directions are re-estimated and $\alpha$ is independently selected on the unchanged validation split for every resample before test evaluation.

\begin{figure}[h]
    \centering
    \includegraphics[width=\linewidth]{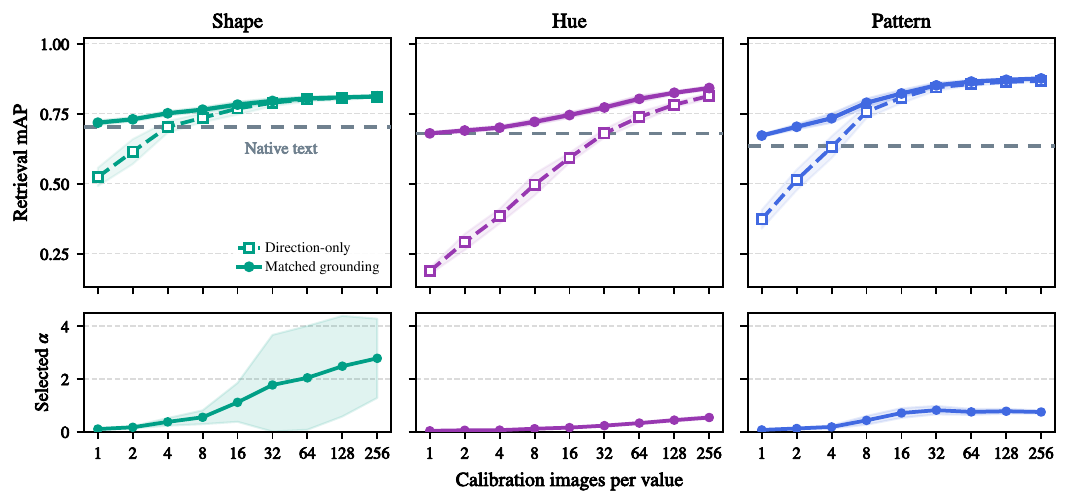}
    \caption{
    \textbf{Finite grounding and the pure visual endpoint under limited calibration support.}
    Top: test mAP for direction-only retrieval and validation-selected finite grounding.
    Bottom: corresponding selected grounding strength $\alpha$.
    Points and shading show mean $\pm 1$ standard deviation over 12 balanced calibration resamples; the dashed line denotes native-text retrieval.
    }
    \label{fig:support_direction_endpoint_alpha}
\end{figure}

As shown in Fig.~\ref{fig:support_direction_endpoint_alpha}, finite grounding and direction-only retrieval differ substantially at small support, particularly for hue and pattern, and the gap generally narrows as support increases.
From one to 256 images per value, direction-only mAP increases from $.524$ to $.811$ for shape, $.187$ to $.813$ for hue, and $.372$ to $.867$ for pattern; the corresponding finite-grounding trajectories are $.718$ to $.812$, $.679$ to $.841$, and $.672$ to $.876$.

The selected $\alpha$ also changes with support.
At one image per value, mean $\alpha$ is $.10$, $.03$, and $.06$ for shape, hue, and pattern; at 256 images per value, it reaches $2.78$, $.53$, and $.74$.
Thus, validation selects only a small movement from the native query at very limited support, with factor-dependent movement along the intervention path as additional calibration evidence becomes available.

For completeness, the resampled sweep stops at 256 images per value so that every plotted point follows the same 12-resample protocol.
Using the complete calibration support, finite/direction-only mAP is $.813/.812$ for shape, $.854/.833$ for hue, and $.878/.870$ for pattern, with selected $\alpha$ values $2.90$, $.65$, and $.70$, respectively.

\paragraph{Scope.}

This analysis characterizes the intervention path rather than attributing retrieval performance to separate linguistic and visual components.
Because both the estimated visual direction and the validation-selected $\alpha$ vary with calibration support, $\alpha$ should not be interpreted as a mixture weight or as the fraction of performance contributed by either component.
The comparison instead shows that finite grounding and the pure visual endpoint need not behave interchangeably across support regimes.

\section{Backbone-, Protocol-, and Value-Resolved Readouts}
\label{app:readouts}

This appendix expands the breadth results summarized in Table~\ref{tab:breadth_full}.
Factor-level mAP is the macro average of one-vs-rest AP over the complete factor vocabulary.
Factor-value top-1 is image-wise classification accuracy obtained by assigning each image to its highest-scoring value query.
It is distinct from the exact-composition retrieval R@1 reported later.

\subsection{Unseen Factor Combinations}

The unseen-combination evaluation withholds combinations of the two non-target factors while keeping every individual value represented in calibration.
Pattern is evaluated on unseen hue--shape cells, hue on unseen pattern--shape cells, and shape on unseen pattern--hue cells.
Visual directions and grounding strengths are re-estimated independently for each backbone and held-out protocol.

\begin{table*}[h]
    \centering
    \scriptsize
    \setlength{\tabcolsep}{3.8pt}
    \renewcommand{\arraystretch}{1.08}
    \caption{
    \textbf{Per-backbone matched access gains on unseen factor combinations.}
    Each entry reports native $\rightarrow$ \textbf{matched}.
    }
    \label{tab:unseen_full}
    \begin{tabular}{l cc cc cc}
        \toprule
        & \multicolumn{2}{c}{Pattern}
        & \multicolumn{2}{c}{Hue}
        & \multicolumn{2}{c}{Shape} \\
        \cmidrule(lr){2-3}
        \cmidrule(lr){4-5}
        \cmidrule(lr){6-7}
        Backbone
        & mAP & Top-1
        & mAP & Top-1
        & mAP & Top-1 \\
        \midrule
        CLIP ViT-B/16
        & $.371 \rightarrow \mathbf{.838}$ & $.434 \rightarrow \mathbf{.856}$
        & $.650 \rightarrow \mathbf{.820}$ & $.579 \rightarrow \mathbf{.856}$
        & $.520 \rightarrow \mathbf{.729}$ & $.549 \rightarrow \mathbf{.707}$ \\

        EVA02-B/16
        & $.533 \rightarrow \mathbf{.856}$ & $.562 \rightarrow \mathbf{.888}$
        & $.578 \rightarrow \mathbf{.816}$ & $.645 \rightarrow \mathbf{.841}$
        & $.579 \rightarrow \mathbf{.817}$ & $.660 \rightarrow \mathbf{.787}$ \\

        FG-CLIP Base
        & $.587 \rightarrow \mathbf{.864}$ & $.529 \rightarrow \mathbf{.910}$
        & $.698 \rightarrow \mathbf{.906}$ & $.710 \rightarrow \mathbf{.950}$
        & $.606 \rightarrow \mathbf{.746}$ & $.602 \rightarrow \mathbf{.736}$ \\

        SigLIP SO400M
        & $.691 \rightarrow \mathbf{.908}$ & $.739 \rightarrow \mathbf{.921}$
        & $.662 \rightarrow \mathbf{.838}$ & $.613 \rightarrow \mathbf{.908}$
        & $.735 \rightarrow \mathbf{.862}$ & $.756 \rightarrow \mathbf{.847}$ \\

        SigLIP2 Base
        & $.673 \rightarrow \mathbf{.895}$ & $.709 \rightarrow \mathbf{.924}$
        & $.672 \rightarrow \mathbf{.872}$ & $.601 \rightarrow \mathbf{.910}$
        & $.710 \rightarrow \mathbf{.832}$ & $.695 \rightarrow \mathbf{.810}$ \\

        SigLIP2 Large
        & $.651 \rightarrow \mathbf{.888}$ & $.669 \rightarrow \mathbf{.924}$
        & $.576 \rightarrow \mathbf{.745}$ & $.570 \rightarrow \mathbf{.766}$
        & $.753 \rightarrow \mathbf{.853}$ & $.788 \rightarrow \mathbf{.846}$ \\

        SigLIP2 SO400M
        & $.649 \rightarrow \mathbf{.884}$ & $.689 \rightarrow \mathbf{.922}$
        & $.491 \rightarrow \mathbf{.613}$ & $.426 \rightarrow \mathbf{.641}$
        & $.769 \rightarrow \mathbf{.861}$ & $.782 \rightarrow \mathbf{.848}$ \\

        \midrule
        \textbf{Macro}
        & $.594 \rightarrow \mathbf{.876}$ & $.619 \rightarrow \mathbf{.906}$
        & $.618 \rightarrow \mathbf{.801}$ & $.592 \rightarrow \mathbf{.839}$
        & $.667 \rightarrow \mathbf{.814}$ & $.690 \rightarrow \mathbf{.798}$ \\
        \bottomrule
    \end{tabular}
\end{table*}

Matched grounding improves both metrics for all 21 target-aligned backbone--factor evaluations.
This result tests generalization to unseen combinations of observed semantic values rather than transfer to unseen nuisance contexts.

\subsection{Balanced Nuisance Partitions}

We construct 24 deterministic balanced partitions of the joint context--renderer realizations.
Each partition assigns eight groups to calibration, four to validation, and twelve to testing.
Visual directions are estimated from calibration data and the factor-level grounding strength is selected on validation independently within every fold.

\begin{table}[h]
    \centering
    \small
    \setlength{\tabcolsep}{5pt}
    \caption{
    \textbf{Matched access gains across 24 balanced nuisance partitions.}
    Values are mean $\pm$ standard deviation across folds after averaging the seven backbones within each fold.
    }
    \label{tab:partition_full}
    \begin{tabular}{lccr}
        \toprule
        Factor
        & mAP: native $\rightarrow$ matched
        & Top-1: native $\rightarrow$ matched
        & Positive endpoints \\
        \midrule
        Pattern
        & $.5701{\pm}.0014 \rightarrow \mathbf{.8618{\pm}.0005}$
        & $.6031{\pm}.0019 \rightarrow \mathbf{.8977{\pm}.0008}$
        & 168/168 \\

        Hue
        & $.6207{\pm}.0026 \rightarrow \mathbf{.7877{\pm}.0032}$
        & $.5961{\pm}.0019 \rightarrow \mathbf{.8182{\pm}.0040}$
        & 168/168 \\

        Shape
        & $.6630{\pm}.0026 \rightarrow \mathbf{.8093{\pm}.0030}$
        & $.6871{\pm}.0031 \rightarrow \mathbf{.7905{\pm}.0049}$
        & 168/168 \\

        \bottomrule
    \end{tabular}
\end{table}

Each factor contributes $7$ backbones $\times$ $24$ folds $=168$ endpoints.
Matched grounding improves both mAP and top-1 at every endpoint.
The small fold-level variation indicates robustness to split choice within the FactorAtlas nuisance generator; it should not be interpreted as transfer to an unrelated renderer distribution.

\subsection{Value-Resolved Access Gains}

Aggregate factor scores can hide whether access gains are shared broadly across the vocabulary or dominated by a few values. 
We therefore recompute AP separately for every one of the 10 pattern, 12 hue, and 8 shape values, and then average each value over the seven backbones. 
Native and matched entries use the same primary context holdout; only the query changes.

Factor-level improvement need not be uniform over individual values.
Table~\ref{tab:value_resolved} reports per-value AP averaged across the seven backbones under the primary context holdout.

\begin{table*}[h]
    \centering
    \scriptsize
    \setlength{\tabcolsep}{3.8pt}
    \renewcommand{\arraystretch}{1.08}
    \caption{
    \textbf{Value-resolved AP under the primary context holdout.}
    Each entry reports the seven-backbone macro native $\rightarrow$ matched AP and its change. 
    Access gains are broad but value-dependent; \textit{teal} is the only value with a negative macro change in this protocol.
    }
    \label{tab:value_resolved}
    \begin{tabular}{lcc@{\hspace{12pt}}lcc@{\hspace{12pt}}lcc}
        \toprule
        \multicolumn{3}{c}{Pattern} &
        \multicolumn{3}{c}{Hue} &
        \multicolumn{3}{c}{Shape} \\
        \cmidrule(lr){1-3}\cmidrule(lr){4-6}\cmidrule(lr){7-9}
        Value & Native $\rightarrow$ Matched & $\Delta$ &
        Value & Native $\rightarrow$ Matched & $\Delta$ &
        Value & Native $\rightarrow$ Matched & $\Delta$ \\
        \midrule
        plain     & $.227\rightarrow.770$ & +.542 & red     & $.825\rightarrow.898$ & +.072 & cube             & $.551\rightarrow.823$ & +.271 \\
        striped   & $.372\rightarrow.854$ & +.482 & orange  & $.789\rightarrow.923$ & +.133 & sphere           & $.895\rightarrow.978$ & +.083 \\
        wavy      & $.592\rightarrow.763$ & +.171 & yellow  & $.767\rightarrow.886$ & +.119 & cylinder         & $.832\rightarrow.908$ & +.076 \\
        checkered & $.975\rightarrow.998$ & +.023 & lime    & $.669\rightarrow.878$ & +.209 & cone             & $.753\rightarrow.915$ & +.161 \\
        dotted    & $.643\rightarrow.884$ & +.241 & green   & $.623\rightarrow.842$ & +.218 & torus            & $.885\rightarrow.999$ & +.114 \\
        cracked   & $.865\rightarrow.996$ & +.131 & teal    & $.606\rightarrow.594$ & -.011 & pyramid          & $.514\rightarrow.649$ & +.135 \\
        mottled   & $.517\rightarrow.920$ & +.403 & cyan    & $.445\rightarrow.619$ & +.174 & triangular prism & $.261\rightarrow.466$ & +.205 \\
        speckled  & $.077\rightarrow.547$ & +.469 & blue    & $.630\rightarrow.897$ & +.267 & octahedron       & $.589\rightarrow.683$ & +.094 \\
        marbled   & $.726\rightarrow.931$ & +.205 & indigo  & $.379\rightarrow.758$ & +.379 &                  &                       &       \\
        brick     & $.677\rightarrow.953$ & +.276 & purple  & $.577\rightarrow.729$ & +.152 &                  &                       &       \\
                  &                       &       & magenta & $.695\rightarrow.852$ & +.157 &                  &                       &       \\
                  &                       &       & pink    & $.427\rightarrow.558$ & +.131 &                  &                       &       \\
        \bottomrule
    \end{tabular}
\end{table*}

The table shows that access gains are distributed across the vocabularies rather than arising from a single value. 
Large gains occur for values such as \textit{plain}, \textit{striped}, \textit{speckled}, \textit{indigo}, and \textit{cube}, while values already near ceiling, such as \textit{checkered} and \textit{torus}, leave less room for improvement.
\textit{Teal} is the sole negative seven-backbone macro endpoint in this protocol. 
We therefore interpret the result as broad factor-level access gains, not as a claim that every individual value must improve.

\section{Global Alignment and Alternative Adaptation}
\label{app:global_alignment_full}

\subsection{Full-Rank Learned Alignment}

The first global comparison asks whether the observed access gap can be removed by learning one expressive transformation over a whole vocabulary, rather than by grounding values individually. 
This baseline receives direct factor supervision and is retrained for every held-out protocol, making it a target-aware broad correction rather than a zero-shot transfer map.

The global baseline applies $q^G=\operatorname{norm}(Aq)$ with a bias-free full-rank map initialized to identity. 
Encoders remain frozen. 
We train one shared map or one map per factor from calibration image--caption pairs, using cyclic factor replacements as hard negatives. 
Training alternates \texttt{a \{pattern\} \{hue\} \{shape\} object} and \texttt{a \{hue\} \{shape\} with a \{pattern\} surface}.
We use AdamW for 30 epochs, batch size 256, weight decay $10^{-5}$, learning rates $\{10^{-4},3\times10^{-4},10^{-3}\}$, and seed 20260820.
Checkpoints are chosen by validation macro-mAP and top-1. 
Each held-out protocol is trained separately. 
Before training, caption-cache rows are asserted against image identifiers and metadata hashes to verify ordering.

\begin{table}[h]
\centering\small
\caption{\textbf{Primary-context global comparison.} 
Seven-backbone macro-mAP.}
\label{tab:global_factor_full}
\begin{tabular}{lcccc}\toprule
Factor&Native&Shared&Factor&Matched\\\midrule
Pattern&.567&.779&.806&\textbf{.862}\\
Hue&.619&.528&.580&\textbf{.786}\\
Shape&.660&.739&.762&\textbf{.803}\\\midrule
Macro&.616&.682&.716&\textbf{.817}\\\bottomrule
\end{tabular}
\end{table}

Table~\ref{tab:global_factor_full} compares four query representations on the same primary-context test galleries. 
Shared alignment uses one map for all three factors, factor alignment uses a separately trained map for each factor, and matched grounding uses the value-specific image contrast without learning a full-rank map. 
The global maps substantially improve pattern and shape. 
Their effect is not uniform, however, and both global maps reduce hue mAP relative to native retrieval in this split. 
Matched grounding has the highest macro result and the highest result for each factor. 
The point is not that global alignment fails, but that a broad correction does not uniformly eliminate value-specific access gaps.

\subsection{Additional Whole-Vocabulary Linear Maps}
\label{app:global_map_families}

We also compare closed-form maps trained from whole-vocabulary centroid correspondences. 
A factor-specific map uses the $K$ native text queries and $K$ calibration image centroids of one factor; a shared map uses all 30 pattern, hue, and shape correspondences. 
Validation selects among ridge $\lambda\in\{10^{-4},10^{-3},10^{-2},10^{-1},1,10,100\}$, truncated low-rank ridge with ranks $\{1,2,4,K-1\}$ (or 29 for the shared map), and row-span orthogonal Procrustes. 
Each candidate is blended with the identity using weights $\{0,.125,\ldots,1\}$. 
Selection uses validation macro-mAP, then top-1, then the smaller map blend.

\begin{table}[h]
\centering\small
\caption{\textbf{Additional whole-vocabulary maps on SigLIP2 Base.} 
Values are primary-context test mAP. 
Validation selects Procrustes for all map scopes in this split.}
\label{tab:centroid_map_families}
\begin{tabular}{lccc}\toprule
Method&Pattern&Hue&Shape\\\midrule
Native&.635&.678&.702\\
Factor-specific map&.772&.727&.789\\
Shared map&.775&.712&.768\\
Matched grounding&\textbf{.878}&\textbf{.854}&\textbf{.813}\\\bottomrule
\end{tabular}
\end{table}

These closed-form baselines test whether the conclusion depends on the optimization details of the full-rank learned map. 
Although validation chooses Procrustes in this particular split, the candidate family also includes ridge and low-rank solutions and is selected without test data. 
Both factor-specific and shared maps improve over native retrieval, but neither reaches matched grounding on any of the three factors in Table~\ref{tab:centroid_map_families}. 
This provides a second, independently fitted whole-vocabulary comparison rather than treating a single global learner as definitive.

\subsection{Post-Global Residual Access Gains}

The direct comparison above still leaves open whether global alignment and matched grounding address the same part of the mismatch. 
We therefore apply them sequentially. For post-global evaluation, the selected global map is first frozen. 
Image directions are then recomputed from calibration data, and a new factor-level $\alpha$ is selected on validation. 
The global map is never updated using the residual result, and test data are used only for the final evaluation. 
A positive gain therefore means that value-matched image-side structure remains useful for improving access after the broad transformation has already been applied.

\begin{table*}[h]
\centering\small
\caption{\textbf{Residual access gains after global alignment.} 
Macro-mAP averages context and three unseen-combination protocols.}
\label{tab:post_global_model_full}
\begin{tabular}{lccc|ccc}\toprule
&\multicolumn{3}{c}{Shared global}&\multicolumn{3}{c}{Factor global}\\
Backbone&Global&+Matched&Improved&Global&+Matched&Improved\\
\midrule
CLIP B/16&.680&.743&11/12&.735&.765&10/12\\
EVA02-B/16&.725&.794&10/12&.738&.791&10/12\\
FG-CLIP Base&.708&.785&10/12&.753&.797&8/12\\
SigLIP SO400M&.764&.823&9/12&.775&.827&9/12\\
SigLIP2 Base&.740&.813&11/12&.753&.816&11/12\\
SigLIP2 Large&.758&.794&7/12&.762&.795&7/12\\
SigLIP2 SO400M&.699&.747&9/12&.717&.752&8/12\\
\midrule
Macro&.725&\textbf{.785}&67/84&.748&\textbf{.792}&63/84\\
\bottomrule
\end{tabular}
\end{table*}

The table averages four held-out conditions, with all three target factors evaluated under each condition, giving 12 endpoints per backbone.
Adding matched grounding raises the shared-global macro from $.725$ to $.785$ and the factor-global macro from $.748$ to $.792$. 
The gain is positive for 67/84 and 63/84 individual endpoints, respectively, so it is strong but not universal. 
In the primary context holdout alone, adding matched grounding yields positive gains for all 21 factor--backbone endpoints: shared rises from $.682$ to $.802$ and factor global from $.716$ to $.806$. 
These residual gains show that the tested broad maps leave residual value-specific access gaps that matched visual grounding can further reduce.

\subsection{Full-Support Cache Adaptation}
\label{app:tip_adapter}

We additionally test whether comparable held-out gains can be obtained by directly retrieving through the labeled calibration support.
We implement a training-free Tip-Adapter-style cache~\citep{zhang2021tip} using all calibration images, while keeping the image and text encoders frozen.
This is a full-support cache control rather than a canonical few-shot reproduction of Tip-Adapter or its fine-tuned variant.

Let $K$ contain the normalized calibration image embeddings and let $L$ contain class-count-normalized factor-wise one-hot labels.
For a normalized test-image embedding $z$, the cache score is
\[
s_{\mathrm{cache}}(z)
=
\exp\!\left(
\beta(zK^\top-1)
\right)L.
\]
The final score combines the native text similarity with the cache score:
\[
s_{\mathrm{Tip}}(z)
=
s_{\mathrm{native}}(z)
+
\eta s_{\mathrm{cache}}(z).
\]
The cache contains all 7,680 calibration images, corresponding to 768 examples per pattern value, 640 per hue value, and 960 per shape value.
The affinity sharpness $\beta$ and mixture strength $\eta$ are selected using validation mAP, with factor-value top-1 used as a tie-breaker.
We also evaluate the cache-only and nearest-support limits to ensure that the selected result is not caused by a truncated hyperparameter range.
All choices are fixed before evaluation on the primary-context test gallery.

\begin{table*}[h]
\centering
\small
\setlength{\tabcolsep}{6pt}
\renewcommand{\arraystretch}{1.08}
\caption{
\textbf{Full-support cache adaptation across backbones and factors.}
Each factor cell reports native $\rightarrow$ Tip-Adapter-style cache $\rightarrow$ \textbf{target-versus-rest grounding} full-gallery mAP under the primary context holdout.
All hyperparameters are selected on validation and fixed before test evaluation.
}
\label{tab:tip_adapter_full}
\begin{tabular}{lccc}
\toprule
Backbone & Pattern & Hue & Shape \\
\midrule
CLIP ViT-B/16
& $.361 \rightarrow .800 \rightarrow \mathbf{.830}$
& $.649 \rightarrow .703 \rightarrow \mathbf{.809}$
& $.519 \rightarrow .677 \rightarrow \mathbf{.723}$ \\

EVA02-B/16
& $.515 \rightarrow .797 \rightarrow \mathbf{.843}$
& $.586 \rightarrow .644 \rightarrow \mathbf{.796}$
& $.563 \rightarrow .740 \rightarrow \mathbf{.804}$ \\

FG-CLIP Base
& $.555 \rightarrow .827 \rightarrow \mathbf{.848}$
& $.678 \rightarrow .767 \rightarrow \mathbf{.892}$
& $.599 \rightarrow .688 \rightarrow \mathbf{.740}$ \\

SigLIP SO400M
& $.663 \rightarrow .847 \rightarrow \mathbf{.891}$
& $.667 \rightarrow .753 \rightarrow \mathbf{.825}$
& $.730 \rightarrow .827 \rightarrow \mathbf{.850}$ \\

SigLIP2 Base
& $.635 \rightarrow .841 \rightarrow \mathbf{.878}$
& $.678 \rightarrow .723 \rightarrow \mathbf{.854}$
& $.702 \rightarrow .760 \rightarrow \mathbf{.813}$ \\

SigLIP2 Large
& $.620 \rightarrow .822 \rightarrow \mathbf{.875}$
& $.590 \rightarrow .657 \rightarrow \mathbf{.736}$
& $.744 \rightarrow .786 \rightarrow \mathbf{.837}$ \\

SigLIP2 SO400M
& $.621 \rightarrow .821 \rightarrow \mathbf{.867}$
& $.488 \rightarrow .560 \rightarrow \mathbf{.592}$
& $.764 \rightarrow .815 \rightarrow \mathbf{.851}$ \\
\midrule
Seven-backbone macro
& $.567 \rightarrow .822 \rightarrow \mathbf{.862}$
& $.619 \rightarrow .687 \rightarrow \mathbf{.786}$
& $.660 \rightarrow .756 \rightarrow \mathbf{.803}$ \\
\bottomrule
\end{tabular}
\end{table*}

The full-support cache yields substantial gains over native retrieval for every backbone and factor.
Target-versus-rest grounding nevertheless achieves higher full-gallery mAP in all 21 backbone--factor cases.
Averaged across all factors and backbones, mAP increases from $.616$ under native retrieval to $.755$ with the Tip-Adapter-style cache and $.817$ with target-versus-rest grounding.
Thus, labeled support can itself yield substantial gains through cache-based adaptation, but the matched target-versus-rest intervention yields larger full-gallery gains without retaining the calibration set as an inference-time cache.

\section{Exact Composition Details}
\label{app:composition_full}

\subsection{Common Factor-Posterior Readout}

Factor-level mAP evaluates one semantic axis at a time and does not show that the exact shape--hue--pattern target outranks near-miss images that differ in one or more factors.
We therefore evaluate exact compositions using the same readout for native and matched queries, isolating the effect of improved factor-level access.

For each context--renderer-seed pair in the evaluation split, we form a separate retrieval gallery.
Under the context holdout, each gallery contains all 960 shape--hue--pattern triples exactly once.
For unseen-pair protocols, evaluation is restricted to triples in the corresponding held-out factor-pair cells.
We average over all valid query--gallery pairs, yielding $6\times2\times960=11{,}520$ instances for the context holdout and $12\times2\times120=2{,}880$ for each unseen-pair protocol.

We use
\begin{equation}
\log p_f(v\mid x)=\frac{\langle q_{f,v},x\rangle}{\tau}
-\log\sum_{u\in\mathcal V_f}\exp\!\left(
\frac{\langle q_{f,u},x\rangle}{\tau}\right),\quad
S(y,x)=\sum_f w_f\log p_f(y_f\mid x).
\end{equation}
Factor-level $\alpha_f$ values are fixed first. We select
$\tau\in\{.005,.01,.02,.05,.1,.2,.5\}$ and positive simplex weights on a
0.1 grid using validation exact MRR followed by R@1. Native and matched use
the same gallery and composition operator.

\begin{table}[h]
\centering\small
\caption{\textbf{Exact composition.} Seven-backbone macro results.}
\label{tab:composition_full}
\begin{tabular}{lccc|ccc}
\toprule
&\multicolumn{3}{c}{Native}&\multicolumn{3}{c}{Matched}\\
Condition&R@1&R@5&mAP&R@1&R@5&mAP\\
\midrule
Context&.357&.714&.517&\textbf{.711}&\textbf{.962}&\textbf{.819}\\
Unseen H$\times$S&.631&.929&.764&\textbf{.928}&\textbf{.998}&\textbf{.960}\\
Unseen P$\times$S&.622&.952&.766&\textbf{.876}&\textbf{.995}&\textbf{.931}\\
Unseen P$\times$H&.661&.937&.779&\textbf{.828}&\textbf{.990}&\textbf{.900}\\
\bottomrule
\end{tabular}
\end{table}

Because each evaluated query has a single exact positive, AP equals the reciprocal rank of that target, so the reported mAP is numerically identical to mean reciprocal rank.
Under the context holdout, matched grounding raises exact R@1 from $.357$ to $.711$ and R@5 from $.714$ to $.962$.
Gains persist for unseen factor-pair combinations: in unseen H$\times$S, for example, pattern queries are evaluated on hue--shape cells withheld from calibration, with P$\times$S and P$\times$H defined analogously.
Thus, stronger factor-level access carries over to exact retrieval of new multi-factor combinations.

\section{Selectivity and Sample Efficiency}
\label{app:selective_grounding_full}

\subsection{Validation-Guided Selective Grounding}

Matched grounding need not be applied to every value.
Some native queries are already effective, while others show a clear validation gain after grounding.
This experiment asks whether held-out audit evidence can select between the native and grounded query for each value without consulting test performance.

We first select one factor-wide grounding strength $\alpha$ on validation.
For each value, we then estimate the uncertainty of its validation AP gain using 2,000 bootstrap resamples over context--renderer-replica groups.
Grounding is retained only when the lower bound of the resulting 95\% confidence interval is positive; otherwise, the native query is kept.
All decisions are fixed before test evaluation.
Here, the unseen H$\times$S protocol uses the same held-out hue--shape cells when evaluating all three target factors; it is distinct from the target-specific unseen-combination macro reported in the main breadth analysis.

\begin{table*}[h]
\centering
\small
\caption{\textbf{Selective grounding on SigLIP2 Base.}}
\label{tab:selective_full}
\begin{tabular}{llcccccc}
\toprule
Protocol&Factor&$\alpha$&Ground/keep&Native&All&Selective&Improved/declined\\
\midrule
Context&Pattern&.70&10/0&.635&.878&.878&10/0\\
&Hue&.65&11/1&.678&.854&.855&11/0\\
&Shape&2.90&7/1&.702&.813&.813&7/0\\
Unseen H$\times$S&Pattern&.70&9/1&.673&.895&.895&9/0\\
&Hue&.10&7/5&.671&.698&.697&6/1\\
&Shape&7.35&7/1&.721&.793&.809&7/0\\\midrule
\multicolumn{3}{l}{Macro/total}&51/9&.6797&.8220&\textbf{.8245}&50/1\\
\bottomrule
\end{tabular}
\end{table*}

Across the two protocols and three factors, the gate grounds 51 of 60 values and retains the native query for nine. 
Relative to native mAP of $.6797$, grounding all values reaches $.8220$ and the validation-guided policy reaches $.8245$. 
Of the 51 grounded decisions, 50 improve on test and one declines.
The small aggregate difference from grounding everything is not the primary point; the result demonstrates that held-out audit evidence can localize where grounding is useful without inspecting test outcomes. 
We treat this as a consequence of the audit rather than an additional central method claim.

\subsection{labeled Support-Size Scaling}
\label{app:support_size}

The main experiments use all available calibration images, but the visual contrast may be estimated from substantially less labeled support.
We vary the number of calibration images per value while keeping the validation and test sets unchanged.
This experiment measures both matched access gain and the agreement between each estimated direction and its full-support counterpart.

For each value, we sample $n\in\{1,2,4,8,16,32,64,128,256\}$ calibration images without replacement and re-estimate its centroid and target-versus-average-rest direction.
Sampling is balanced across values, and each finite support budget is evaluated using 12 independent calibration resamples.
For every resample, the grounding strength $\alpha$ is selected again on the unchanged validation split.
The test set is never sampled or used for selection.
One-shot support therefore uses one image per value, corresponding to 10, 12, or 8 total calibration images for pattern, hue, or shape, respectively.

Table~\ref{tab:support_size} reports results for SigLIP2 Base, averaged across the three factors and two held-out protocols, yielding six factor--protocol endpoints.
The first protocol is the primary context holdout.
The second applies the same held-out hue--shape cells when evaluating all three target factors.
Direction cosine measures the cosine similarity between each support-limited direction and the corresponding direction estimated from all calibration images.

\begin{table*}[h]
\centering
\small
\setlength{\tabcolsep}{4.2pt}
\caption{\textbf{Matched access gains with increasing labeled support.} 
Finite-budget results average 12 calibration resamples.}
\label{tab:support_size}
\begin{tabular}{lrrrrrrrrrr}
\toprule
Images/value&1&2&4&8&16&32&64&128&256&Full\\
\midrule
Test mAP&.695&.711&.730&.755&.774&.795&.806&.812&.819&.822\\
Gain&+.015&+.032&+.051&+.076&+.095&+.115&+.127&+.133&+.139&+.142\\
Direction cosine&.466&.594&.719&.825&.897&.946&.972&.986&.994&1.000\\
\bottomrule
\end{tabular}
\end{table*}

\begin{table}[h]
\centering
\small
\caption{\textbf{Factor-resolved support-size subset.} Entries are test mAP.}
\label{tab:support_size_factors}
\begin{tabular}{lccccc}
\toprule
Factor&1&8&32&256&Full\\
\midrule
Pattern&.688&.802&.860&.884&.887\\
Hue&.670&.698&.731&.769&.776\\
Shape&.728&.767&.793&.805&.803\\
\bottomrule
\end{tabular}
\end{table}

The first table aggregates the six factor--protocol endpoints, whereas Table~\ref{tab:support_size_factors} separates the three factors after averaging the two protocols. Access gains and direction quality increase smoothly with support size.
Thirty-two images per value achieve about 81\% of the full-support gain, and 64 achieve about 89\%. 
Shape is already close to its full-support mAP at moderate budgets, whereas pattern benefits more strongly from additional support. 
The positive one-shot mean does not imply a stable one-shot direction: its mean cosine to the full-support direction is only $.466$.
Thus the sweep shows that substantial access gains can be obtained with limited labeled support at the aggregate level, while also showing why the full calibration set is preferable for a stable geometric estimate.

\section{Natural-Image Protocols and Full Results}
\label{app:natural_full}

\subsection{Shared Evaluation Protocol}

All natural-image experiments use frozen, normalized image and text embeddings.
For each target value, we estimate normalized calibration centroids and construct a normalized target-versus-average-rest visual direction.
A single factor-level grounding strength is selected on validation by full-gallery macro-mAP and then fixed before test evaluation.
Test images are never used to select prompts, directions, grounding strengths, thresholds, or checkpoints.

Because the natural-image datasets differ in their annotations and grouping structure, we construct dataset-specific calibration, validation, and test partitions while preserving this shared evaluation contract.
The primary metric is full-gallery macro-mAP.

\subsection{Backbone-Resolved Results}

Table~\ref{tab:natural_transfer_full} reports every backbone-level result behind the seven-backbone macro results in the main paper. 
The columns intentionally retain separate datasets and targets rather than pooling incompatible natural protocols: DTD tests texture classes, Fashionpedia tests garment pattern and length, COCO-Facet tests localized material, and UT-Zappos tests footwear material under product-disjoint splits. 
Each cell reports native retrieval followed by matched visual grounding; the final row averages backbones within each setting, not examples across datasets.

\begin{table*}[h]
    \centering
    \small
    \setlength{\tabcolsep}{4.5pt}
    \renewcommand{\arraystretch}{1.12}
    \caption{
    \textbf{Backbone-resolved natural-image transfer.}
    Each cell reports native $\rightarrow$ \textbf{matched} full-gallery macro-mAP.
    Fashionpedia pattern uses the garment-category holdout, and COCO-Facet material uses the natural localized crop.
    Matched grounding improves all 28 backbone--setting endpoints in the four stable transfer settings; UT-Zappos is reported separately as a higher-variance boundary.
    }
    \label{tab:natural_transfer_full}
    \begin{tabular}{lccccc}
        \toprule
        Backbone
        & DTD
        & \multicolumn{2}{c}{Fashionpedia}
        & COCO-Facet
        & UT-Zappos \\
        \cmidrule(lr){3-4}
        & Texture
        & Pattern
        & Length
        & Material
        & Material \\
        \midrule

        CLIP ViT-B/16
        & .359 $\rightarrow$ \textbf{.668}
        & .452 $\rightarrow$ \textbf{.618}
        & .178 $\rightarrow$ \textbf{.317}
        & .604 $\rightarrow$ \textbf{.850}
        & .522 $\rightarrow$ \textbf{.617} \\

        EVA02-B/16
        & .444 $\rightarrow$ \textbf{.731}
        & .567 $\rightarrow$ \textbf{.768}
        & .220 $\rightarrow$ \textbf{.361}
        & .668 $\rightarrow$ \textbf{.819}
        & .665 $\rightarrow$ .665 \\

        FG-CLIP Base
        & .500 $\rightarrow$ \textbf{.751}
        & .566 $\rightarrow$ \textbf{.789}
        & .241 $\rightarrow$ \textbf{.352}
        & .601 $\rightarrow$ \textbf{.899}
        & .631 $\rightarrow$ \textbf{.676} \\

        SigLIP SO400M
        & .621 $\rightarrow$ \textbf{.803}
        & .631 $\rightarrow$ \textbf{.799}
        & .297 $\rightarrow$ \textbf{.396}
        & .615 $\rightarrow$ \textbf{.869}
        & .677 $\rightarrow$ \textbf{.751} \\

        SigLIP2 Base
        & .564 $\rightarrow$ \textbf{.771}
        & .620 $\rightarrow$ \textbf{.789}
        & .285 $\rightarrow$ \textbf{.370}
        & .627 $\rightarrow$ \textbf{.874}
        & .678 $\rightarrow$ \textbf{.717} \\

        SigLIP2 Large
        & .588 $\rightarrow$ \textbf{.797}
        & .620 $\rightarrow$ \textbf{.778}
        & .308 $\rightarrow$ \textbf{.403}
        & .628 $\rightarrow$ \textbf{.860}
        & .662 $\rightarrow$ \textbf{.726} \\

        SigLIP2 SO400M
        & .607 $\rightarrow$ \textbf{.798}
        & .614 $\rightarrow$ \textbf{.793}
        & .302 $\rightarrow$ \textbf{.387}
        & .623 $\rightarrow$ \textbf{.851}
        & .642 $\rightarrow$ \textbf{.724} \\

        \midrule
        \textbf{Seven-backbone macro}
        & .526 $\rightarrow$ \textbf{.760}
        & .581 $\rightarrow$ \textbf{.762}
        & .262 $\rightarrow$ \textbf{.369}
        & .624 $\rightarrow$ \textbf{.860}
        & .639 $\rightarrow$ \textbf{.697} \\

        \bottomrule
    \end{tabular}
\end{table*}

Matched visual grounding yields positive gains for every evaluated backbone on DTD, Fashionpedia pattern, Fashionpedia length, and COCO-Facet material.
It yields positive gains for six of seven backbones on UT-Zappos.
The remaining EVA02-B/16 endpoint changes from $0.6653$ to $0.6649$ and is therefore unchanged at the three-decimal precision used in the paper.
Because the 96-product UT-Zappos cohort also exhibits substantial variation across its five product-disjoint rotations, we treat it as a boundary result rather than part of the stable-transfer count.

\subsection{Dataset-Specific Protocols}

\paragraph{DTD texture.}

We use all 5,640 DTD images and all 47 official texture classes.
Each of the ten official partitions provides 40 calibration, 40 validation, and 40 test images per class.
The official class token is used as the fixed native query.
Directions and grounding strengths are re-estimated independently within each partition, and the table reports performance averaged over the ten partitions.

\paragraph{Fashionpedia garment pattern.}

We use localized garment crops and fine-grained attributes from Fashionpedia.
Each annotated garment box is placed on an aspect-ratio-preserving square canvas and resized to the model input resolution.
The pattern vocabulary contains six values: \textit{plain}, \textit{floral}, \textit{stripe}, \textit{check}, \textit{abstract}, and \textit{dot}.

The resulting cohort contains 2,400 crops, balanced over the Cartesian product of six patterns and four garment categories.
We perform four leave-one-category-out evaluations.
For each run, one garment category supplies the entire test gallery and is excluded from both direction estimation and grounding-strength selection.
This tests whether pattern directions estimated from other garment categories transfer to a previously unseen garment category.

\paragraph{Fashionpedia garment length.}

We separately evaluate an eight-value garment-length vocabulary: \textit{mini}, \textit{micro}, \textit{above-the-knee}, \textit{knee}, \textit{midi}, \textit{below-the-knee}, \textit{maxi}, and \textit{floor}.
The cohort contains 2,847 localized dress crops.

Five source-image-disjoint calibration--validation rotations are constructed from the Fashionpedia training set.
The untouched Fashionpedia validation gallery is used as test in every rotation.
Consequently, variation across rotations measures robustness to the calibration and validation samples rather than uncertainty from different test galleries.

During cohort construction, local English descriptions from
LOTS/Sketchy~\citep{girella2025lots} were linked to the Fashionpedia examples.
However, the evaluated images, garment boxes, category labels, and visual attributes come from Fashionpedia, and retrieval uses fixed factor prompts rather than the LOTS sketch-generation task.
We therefore refer to the evaluated dataset as Fashionpedia while recording both sources in the data provenance.

\paragraph{COCO-Facet material.}

We join official COCO-Facet material labels with COCO-Stuff localization~\citep{caesar2018coco}.
We retain images containing exactly one requested material label and exclude images containing a segment from another requested material class.

For each image, the natural crop is the tight bounding box of the largest individual target-material segment.
It is not the union of spatially disconnected segments.
We apply fixed inclusion criteria requiring the target mask to occupy at least $1\%$ of the image, fill at least $25\%$ of its bounding box, and have a box covering at most $50\%$ of the full image.
We additionally require at least 24 images per material class. 

The resulting cohort contains 306 images: 172 \textit{metal}, 29 \textit{stone}, and 105 \textit{wood}.
Five deterministic rotations are image-disjoint across calibration, validation, and test.
The main-paper result uses the unmodified natural crop.

\paragraph{UT-Zappos material.}

We evaluate four footwear materials: \textit{Leather}, \textit{Suede}, \textit{Patent Leather}, and \textit{Synthetic}.
The cohort is restricted to a fixed footwear stratum and balances material counts within retained toe-style--heel-height cells.

We retain one deterministic image per product, yielding 96 products with 24 products per material.
Products never cross calibration, validation, and test roles.
Results are averaged over five product-disjoint rotations.

\subsection{Cross-Dataset Reuse of Shared Visual Distinctions}
\label{app:cross_dataset_reuse}

The matched natural-image experiments above estimate visual contrasts and select grounding strength separately within each target dataset.
We additionally ask a stricter question: when two domains share the same visual distinction, can image-side directions estimated in one domain remain useful in the other without any target-side fitting or validation?
We test this using the four pattern values shared by FactorAtlas and Fashionpedia.

For each backbone, we construct FactorAtlas pattern directions from the complete ten-value vocabulary using the four calibration contexts and reuse the corresponding grounding strength selected in FactorAtlas.
No Fashionpedia image is used to estimate a direction or select the grounding strength.
We map FactorAtlas \textit{plain}, \textit{striped}, \textit{checkered}, and \textit{dotted} to Fashionpedia \textit{plain}, \textit{stripe}, \textit{check}, and \textit{dot}, respectively.

We evaluate all four Fashionpedia category-held-out partitions.
Each held-out gallery contains 600 images spanning all six Fashionpedia pattern labels, so the two unmapped labels remain natural negatives rather than being removed from the gallery.
Across seven backbones, directly reusing the FactorAtlas directions improves macro-mAP from $.583$ to $.719$ (Table~\ref{tab:cross_dataset_reuse}).
The aggregate gain is positive for all seven backbones and for all 28 backbone--partition evaluations.
At the individual backbone--value level, 20 of 28 cases improve, indicating that cross-dataset reuse is consistent in aggregate but remains value-dependent.
These results suggest that when the same visual distinction is shared across domains, its image-side direction can remain useful beyond the data from which it was estimated.

\begin{table}[h]
\centering
\small
\caption{
\textbf{Cross-dataset reuse of shared pattern distinctions.}
For each backbone, FactorAtlas-derived pattern directions and its fixed FactorAtlas grounding strength are reused without Fashionpedia fitting or validation.
Values are four-value macro-mAP averaged over the four Fashionpedia category-held-out partitions.
}
\label{tab:cross_dataset_reuse}
\setlength{\tabcolsep}{5pt}
\begin{tabular}{lrrr}
\toprule
Backbone & Native & Reused direction & $\Delta$ \\
\midrule
CLIP ViT-B/16      & .414 & \textbf{.490} & +.075 \\
EVA02-B/16         & .575 & \textbf{.720} & +.145 \\
FG-CLIP Base       & .568 & \textbf{.800} & +.232 \\
SigLIP SO400M      & .640 & \textbf{.767} & +.127 \\
SigLIP2 Base       & .631 & \textbf{.795} & +.164 \\
SigLIP2 Large      & .628 & \textbf{.750} & +.121 \\
SigLIP2 SO400M     & .623 & \textbf{.711} & +.088 \\
\midrule
Seven-backbone macro & .583 & \textbf{.719} & +.136 \\
\bottomrule
\end{tabular}
\end{table}

\end{document}